%% file: latex/acl_latex.tex
\documentclass[11pt]{article}

\usepackage[final]{acl}

\usepackage{times}
\usepackage{latexsym}

\usepackage[T1]{fontenc}

\usepackage[utf8]{inputenc}

\usepackage{microtype}

\usepackage{inconsolata}

\usepackage{graphicx}

\usepackage{amssymb}
\usepackage{amsmath} 
\usepackage{booktabs}
\usepackage{multirow}
\usepackage{multicol}
\usepackage{enumerate}
\usepackage{enumitem}
\usepackage{graphicx}

\newcommand{\ourmlong}{Bias \textbf{Fi}tting to \textbf{Mi}tigate Length Bias of \textbf{R}eward \textbf{M}odel}
\newcommand{\ourm}{FiMi-RM}

\title{Bias Fitting to Mitigate Length Bias of Reward Model in RLHF}

\author{
  \textbf{Kangwen Zhao\textsuperscript{1}},
  \textbf{Jianfeng Cai\textsuperscript{1}},
  \textbf{Jinhua Zhu\textsuperscript{1}}
\\
  \textbf{Ruopei Sun\textsuperscript{1}},
  \textbf{Dongyun Xue\textsuperscript{1}},
  \textbf{Wengang Zhou\textsuperscript{1}\thanks{Corresponding author.}},
  \textbf{Li Li\textsuperscript{1}},
  \textbf{Houqiang Li\textsuperscript{1}\footnotemark[\value{footnote}]}
\\
  \textsuperscript{1}University of Science and Technology of China
\\
  \texttt{\{zkwzkw,xiaobaicai,teslazhu\}@mail.ustc.edu.cn}
  \\
  \texttt{\{ruopeisun,andyxue\}@mail.ustc.edu.cn,\{zhwg,lil1,lihq\}@ustc.edu.cn}
}

\begin{document}
\maketitle
\begin{abstract}
Reinforcement Learning from Human Feedback (RLHF) relies on reward models to align large language models with human preferences. However, RLHF often suffers from reward hacking, wherein policy learning exploits flaws in the trained reward model to maximize reward scores without genuinely aligning with human preferences. A significant example of such reward hacking is length bias, where reward models usually favor longer responses irrespective of actual response quality. Previous works on tackling length bias have notable limitations, these approaches either mitigate bias without characterizing the bias form, or simply assume a linear length-reward relation. To accurately model the intricate nature of length bias and facilitate more effective bias mitigation, we propose \ourm{} (\ourmlong{}), a framework that autonomously learns and corrects underlying bias patterns. Our approach consists of three stages: First, we warm up by training a standard reward model which inherently contains length bias. Next, we deploy a lightweight fitting model to capture the non-linear relation between length and reward. Finally, we incorporate this learned relation into the reward model, effectively decoupling length from reward while preserving preference modeling capabilities. Experimental results demonstrate that \ourm{} achieves a more balanced length-reward distribution. Furthermore, when applied to alignment algorithms such as Direct Preference Optimization (DPO) and Best-of-N (BoN), our debiased reward model improves length-controlled win rate and reduces verbosity without compromising its performance.
\end{abstract}
\input{latex/1_Intro}

\input{latex/2_RelatedWorks}

\input{latex/3_Method}
\input{latex/4_Exp}

\input{latex/5_conclusion}

\section*{Acknowledgement}
This work was supported by the National Natural Science Foundation of China under Contract 623B2097, the Youth Innovation Promotion Association CAS. It was also supported by the GPU cluster built by MCC Lab of USTC \& the Supercomputing Center of USTC.
\section*{Limitations}
Here we focus on length debiasing in reward models. Although we have achieved better results in debiasing, whether human preferences are entirely independent of length (or in terms of correlation, whether the Pearson correlation coefficient is truly zero) remains a question worthy of further investigation. From a practical standpoint, empirical observations suggest that humans often favor more detailed responses, which naturally tend to be longer. For instance, in tasks like summarization or open-ended question answering, thorough explanations with supporting evidence are typically rated higher than brief, vague answers; or to put it another way, sometimes users explicitly include requests for longer and more detailed responses in their instructions. These introduce a potential positive correlation between length and preferences.

\bibliography{custom}
\clearpage
\appendix

\begin{figure*}[ht!]
  
  \centering
  \includegraphics[width=1\textwidth]{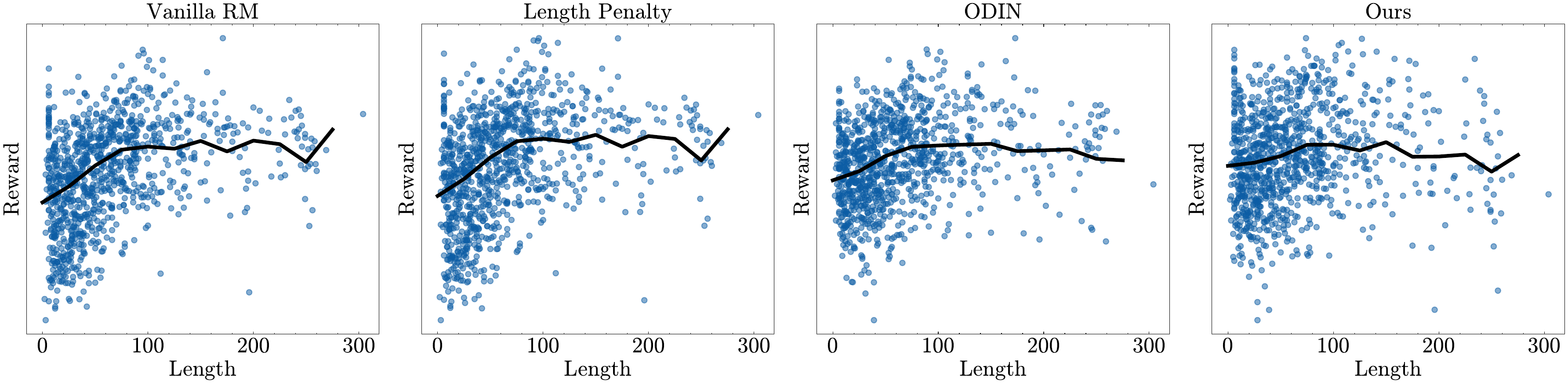}
  \caption{The scatter plot of reward-length relation (Gemma2-9B).}

  \label{fig:scatter_gemma}
\end{figure*}

\begin{figure*}[ht!]
  
  \centering
  \includegraphics[width=1\textwidth]{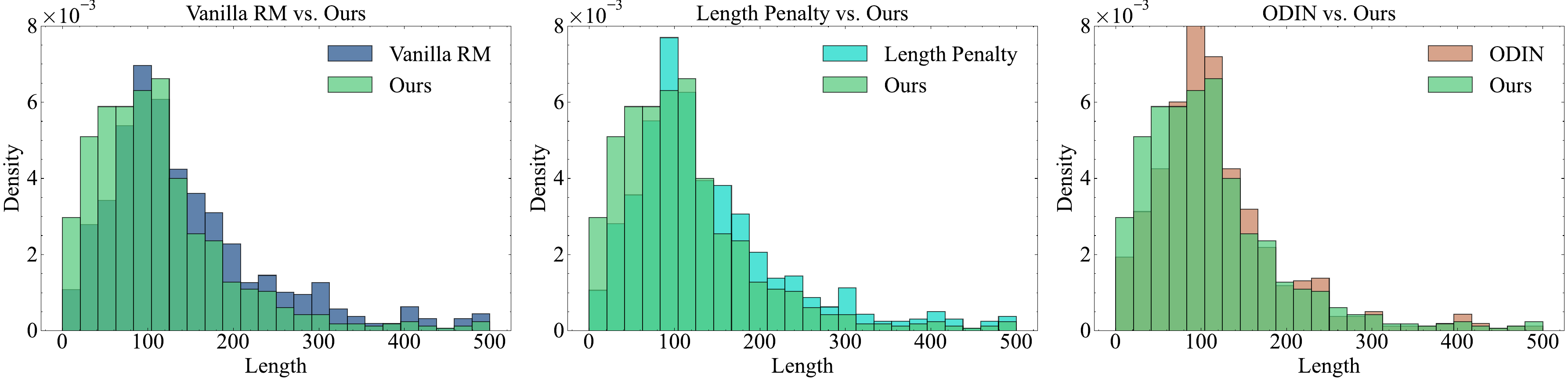}
  \caption{Comparison of the distribution of responses selected by BoN (Gemma2-9B).}

  \label{fig:bon_gemma}
\end{figure*}

\begin{figure*}[ht!]
  
  \centering
  \includegraphics[width=1\textwidth]{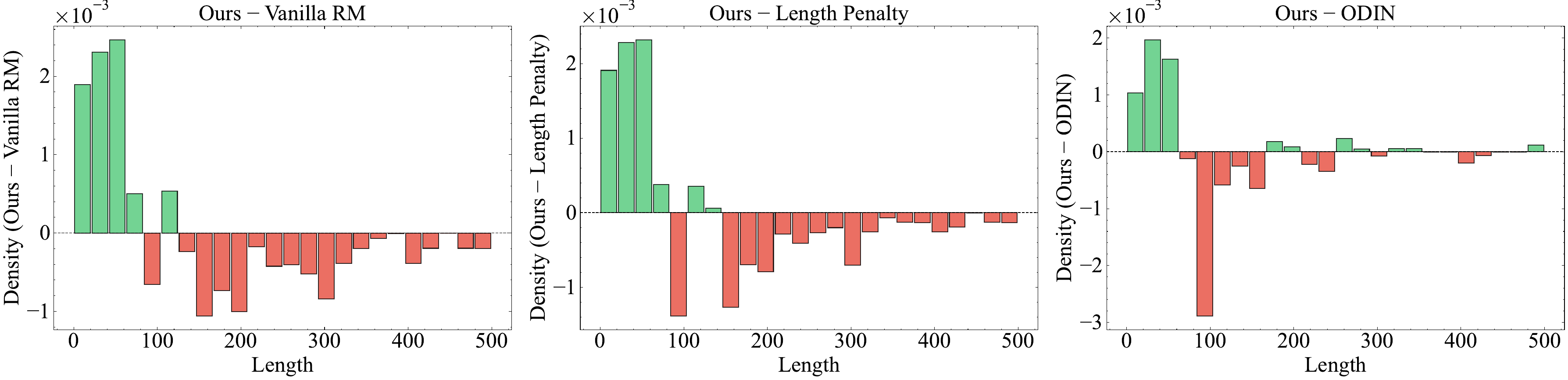}
  \caption{Difference in response length distributions (Gemma2-9B).}

  \label{fig:bon_gemma_div}
\end{figure*}

\begin{figure*}[ht!]
  
  \centering
  \includegraphics[width=1\textwidth]{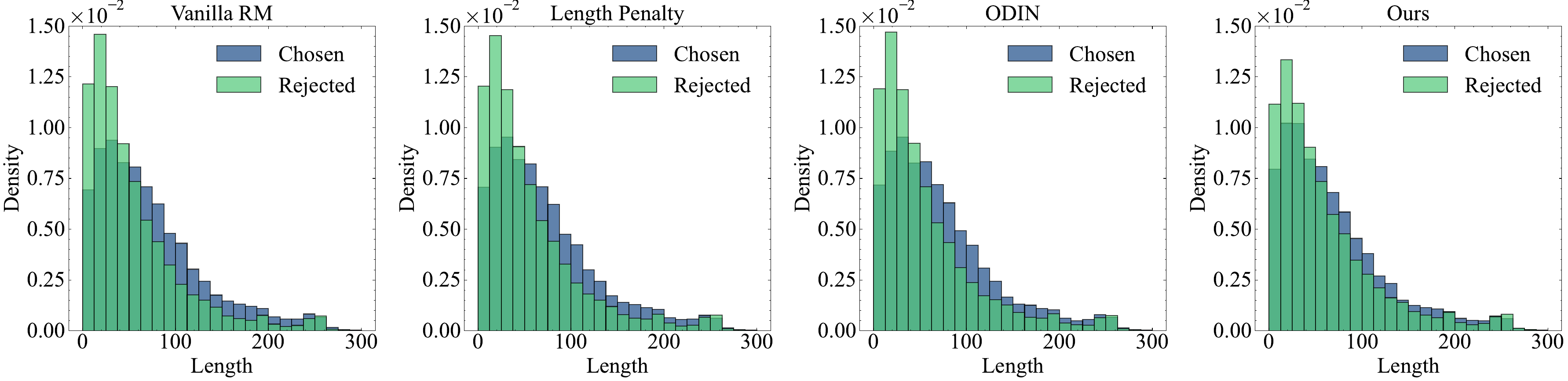}
  \caption{The length distribution of chosen and rejected responses in the labeling stage of DPO (Gemma2-9B).}

  \label{fig:dpo_gemma}
\end{figure*}
\begin{figure*}[ht!]
  
  \centering
  \includegraphics[width=1\textwidth]{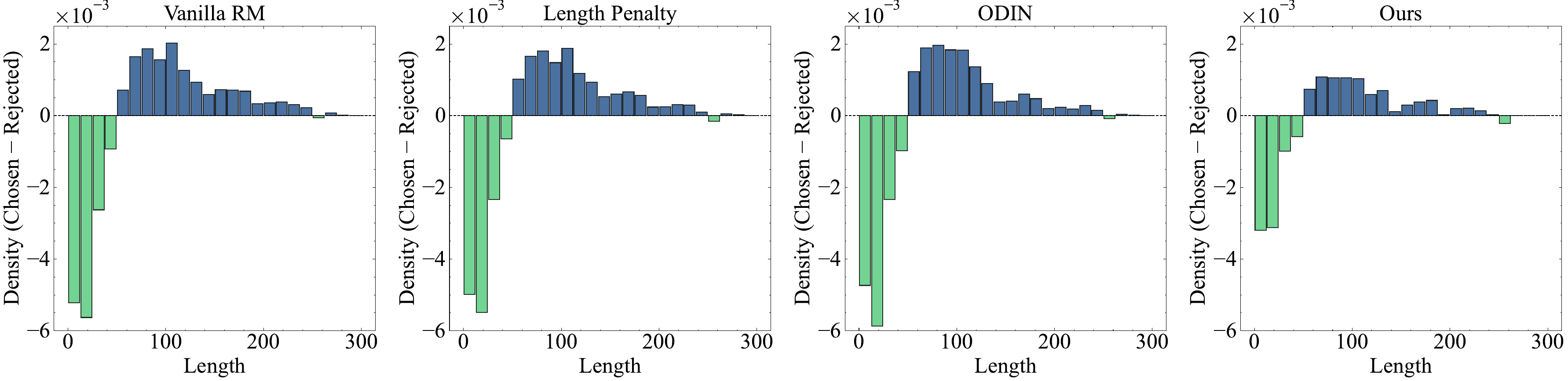}
  \caption{The difference between the chosen and rejected response length distributions (chosen - rejected) shown in Figure \ref{fig:dpo_gemma} (Gemma2-9B).}

  \label{fig:dpo_gemma_div}
\end{figure*}

\section{Ablation Study}
\label{app:ablation}

We conduct ablation experiments with three dimensions: (1) the multi-stage training design vs.\ joint training,
(2) the alternating training schedule in Stage~3,
(3) the architecture of the fitting model $model_f$.
All experiments use Qwen2.5-7B as the base model
and BoN ($N=8$) as the alignment algorithm, evaluated on AlpacaEval.

\paragraph{Multi-Stage Design vs.\ Joint Training.}
Here we explored a joint training strategy where
the fitting model, reward model, and debiasing objective are optimized
together.
While joint training is feasible, its performance is weaker than the
multi-stage design, as shown in Table~\ref{tab:ablation_full}.
By separating bias modeling (Stage~2) and bias mitigation (Stage~3)
into distinct stages, the pipeline allows the fitting model to first
accurately characterize the length bias before the reward model attempts
to remove it.

\paragraph{Alternating Training Schedule}
We compare FiMi-RM against a variant that trains only the reward model
in Stage~3 (without alternating updates to $model_f$).
As shown in Table~\ref{tab:ablation_full}, removing alternating training
degrades both LC-WR and WR, confirming that refining the two
models is beneficial for accurate bias capture and removal.

\paragraph{Architecture of Fitting Model}
We compare our ResNet-based $model_f$ against four alternatives: a simple linear function $f(x)=ax+b$,
Polynomial Regression, 1D-CNN, and MLP.
Results are summarized in Table~\ref{tab:ablation_full}.
The ResNet architecture achieves the best performance.

\begin{table*}[ht]
\centering

\begin{tabular}{lccc}
\toprule
Method & LC-WR$\uparrow$ & WR$\uparrow$ & $L_{\mathrm{char}}$ \\
\midrule
Vanilla RM                              & 69.25 & 73.84 & 638 \\
\midrule
FiMi-RM with Joint Training            & 71.51  & 76.01    & 568  \\
FiMi-RM without Alternating Training   & 71.16 & 74.27 & 588 \\
\midrule
FiMi-RM with Polynomial Regression     & 67.11 & 71.54 & 722 \\
FiMi-RM with Linear Fitting            & 71.67 & 75.30 & 501 \\
FiMi-RM with 1D-CNN                    & 71.15 & 74.29 & 516 \\
FiMi-RM with MLP                       & 71.35 & 75.26 & 550 \\

\midrule

\textbf{FiMi-RM (Ours)}                & \textbf{72.59} & \textbf{76.39} & 543 \\
\bottomrule
\end{tabular}
\caption{
  Ablation results.
  ``Joint Training'' optimizes all objectives together in a single stage.
  ``without Alternating Training'' trains only the reward model in Stage~3.
  ``Linear Fitting'' replaces $model_f$ with $f(x)=ax+b$.
}
\label{tab:ablation_full}
\end{table*}

\section{Visualization Results of Gemma Model}
In the experimental section, we present various visualization results on the Qwen model, and in this appendix, we further provide corresponding results on the Gemma2-9B. Including:
the scatter plot of reward-length (Figure \ref{fig:scatter_gemma}); the comparison of the distribution of responses selected by BoN (Figure \ref{fig:bon_gemma} and \ref{fig:bon_gemma_div}); the length distribution of chosen and rejected responses in the data reannotating stage of DPO (Figure \ref{fig:dpo_gemma} and \ref{fig:dpo_gemma_div}).

From these figures, we observe that although the model family changes, the conclusions remain largely consistent with those obtained earlier on Qwen2.5-7B: our reward model exhibits a smaller bias with respect to length, demonstrating the generality of our approach.

\section{Detailed Network Structure of Fitting Model}
Here we present the detailed structure of our $model_f$ in Table \ref{fig:fit}. As mentioned in method section, the formula for Length Encoding is
\begin{equation}
\scalebox{0.85}{$
\text{LE}(\text{len}(y)) = \Bigg[ \sin\left(\frac{\text{len}(y)}{10000^{2j/d}}\right), \cos\left(\frac{\text{len}(y)}{10000^{2j/d}}\right) \Bigg]_{j=0}^{\frac{d}{2}-1}.$}
\end{equation}
Where the hidden dimension $d$ is set to 32 in experiments. This yields a total of $\approx6.4\text{K}$ parameters for $model_f$, which is negligible compared to the reward models’ sizes (7B and 9B). Since the computation of the Pearson correlation coefficient becomes more accurate with larger batch size $B$, we adopt a multi-GPU aggregation framework, aggregating the batch data across $8$ devices to obtain the length-reward pairs, this effectively increase the value of~$B$. Additionally, due to the relatively small model size, we set the learning rate to $5\text{e-}3$ during training.
\begin{table}[!ht]

\centering
\small
\resizebox{\linewidth}{!}{
\begin{tabular}{l l l}
\toprule
component & layer & input $\to$ output \\
\midrule
embedding
 & Length Encoding
 & $len(y)$:[B] $\to$ [B, d] \\
\midrule
block 1
 & LayerNorm(d) 
 & [B, d] $\to$ [B, d] \\
 & MLP(d$\to$d$\to$d$\to$d) 
 & [B, d] $\to$ [B, d] \\
 & ReLU + residual 
 & [B, d] $\to$ [B, d] \\
\midrule
block 2
 & LayerNorm(d) 
 & [B, d] $\to$ [B, d] \\
 & MLP(d$\to$d$\to$d$\to$d) 
 & [B, d] $\to$ [B, d] \\
 & ReLU + residual 
 & [B, d] $\to$ [B, d] \\
\midrule
head
 & Linear(d$\to$1) 
 & [B, d] $\to$ [B, 1] \\
 & squeeze(1) 
 & [B, 1] $\to$ $\hat{r}:$[B] \\
\bottomrule
\end{tabular}
}
\caption{Structure of $model_f$}
\label{fig:fit}
\end{table}

\section{Computational Cost Analysis}

\begin{table*}[!ht]
\centering
\begin{tabular}{lc}
\toprule
Training Step Type & Time (s/step) \\
\midrule
Standard RM update (Vanilla RM baseline) & $\approx$2.48 \\
Fitting model update (forward through both; backward through $model_f$) & $\approx$0.60 \\
Debiased RM update (forward through both; backward through RM) & $\approx$2.50 \\
\midrule
\textbf{Vanilla RM total} & \textbf{$\approx$43 min/epoch} \\
\textbf{FiMi-RM total}    & \textbf{$\approx$32 min/epoch} \\
\bottomrule
\end{tabular}
\caption{Training time on $8\times$A100 GPUs (Qwen2.5-7B). The fitting model (6.4k parameters) is negligibly small relative to Qwen2.5-7B (7B parameters), so its backward is much more faster. The total time includes not only training time but also other overhead, so it will be longer than simply training time × number of steps.}
\label{tab:compute}
\end{table*}
FiMi-RM introduces a secondary fitting model, but this does not increase total training cost.
To ensure a fair comparison, samples used to update the fitting model are not reused to update the reward model within the same iteration; consequently, the total number of training steps is identical to the Vanilla RM baseline.

Table~\ref{tab:compute} reports average per-step times on $8\times$A100 GPUs, across all 938 steps, FiMi-RM allocates 480 steps to first stage, 240 steps to Stage~2, and the remaining 218 steps to Stage~3.
Since all steps in Stage~2 and half of the steps in Stage~3 are allocated to the faster fitting-model updates, the overall training time per epoch is reduced by approximately 25\% relative to the Vanilla RM.
FiMi-RM therefore achieves length debiasing with no additional computational overhead.

\section{BT Loss Monitoring During Stage~3}
\label{app:bt_loss}

To assess whether the preference modeling capacity of the reward model is maintained during Stage~3, we monitor the Bradley--Terry loss $\mathcal{L}_\textit{BT}$ throughout debiasing training.
On the Dahoas-rm-static dataset with Qwen2.5-7B, the BT loss increases from $0.530$ to $0.566$ over Stage~3.
This rise is an expected consequence of removing the length shortcut: because approximately 58\% of chosen responses are longer, the vanilla model exploits length as a proxy signal, yielding an artificially low BT loss.
By enforcing semantic focus and decoupling length from reward, the training objective becomes harder, which naturally raises the loss.
Critically, consistent improvements across downstream benchmarks confirm that preference modeling ability is preserved and indeed strengthened after debiasing.

\section{Results under SimPO}
\label{app:simpo}

To assess the generality of FiMi-RM beyond DPO and BoN, we conduct the experiments replacing DPO with SimPO~\citep{NEURIPS2024_e099c1c9} (learning rate $5\text{e-}7$, $\beta=2$, $\gamma=0.8$, per-device batch size $2$ and gradient accumulation steps $8$ on Qwen2.5-7B).
As shown in Table~\ref{tab:simpo}, FiMi-RM reduces response length and improves LC-WR compared to the vanilla reward model under SimPO training as well.

\begin{table}[ht]
\centering

\begin{tabular}{lccc}
\toprule
SimPO & LC-WR$\uparrow$ & WR$\uparrow$ & $\text{L}_{char}$ \\
\midrule
Vanilla RM & 53.68 & 58.69 & 711 \\
FiMi-RM   & \textbf{55.03} & \textbf{58.81} & 549 \\
\bottomrule
\end{tabular}
\caption{AlpacaEval results under SimPO training (Qwen2.5-7B).}
\label{tab:simpo}
\end{table}

\section{Evaluation on RLHFlow Dataset}
\label{app:rlhflow}

To evaluate robustness across diverse preference distributions, we train FiMi-RM on the RLHFlow~\citep{dong2024rlhfworkflowrewardmodeling} dataset, a comprehensive dataset with eight sources.
Table~\ref{tab:rlhflow} reports AlpacaEval results using Qwen2.5-7B and BoN ($N=8$).

\begin{table}[ht]
\centering

\begin{tabular}{lccc}
\toprule
Method & LC-WR$\uparrow$ & WR$\uparrow$ & $\text{L}_{char}$ \\
\midrule
Vanilla RM & 62.00 & 68.73 & 1627 \\
FiMi-RM (Ours) & \textbf{66.59} & \textbf{71.72} & 1595 \\
\bottomrule
\end{tabular}
\caption{AlpacaEval results on the RLHFlow dataset (Qwen2.5-7B, BoN).}
\label{tab:rlhflow}
\end{table}

FiMi-RM consistently outperforms the vanilla reward model on this more diverse dataset, achieving higher alignment accuracy with shorter outputs.

\section{Evaluation on other RLHF Setting}
\label{app:grpo}

We futher evaluate our debiased reward model directly in other RLHF setting
using GRPO~\citep{shao2024deepseekmath}.
Table~\ref{tab:grpo} reports AlpacaEval results on Qwen2.5-7B, with learning rate $2\text{e-}6$, batch size $8$ (per-device batch size $1$ with gradient accumula-
tion steps $8$), and sample size $n=8$. FiMi-RM achieves higher LC-WR while producing more concise
responses.

\begin{table}[ht!]
\centering

\begin{tabular}{lccc}
\toprule
GRPO & LC-WR$\uparrow$ & WR$\uparrow$ &  $\text{L}_{char}$ \\
\midrule
Vanilla RM& 58.80 & 72.73 & 1285 \\
FiMi-RM & \textbf{67.42} & \textbf{75.91} & 999 \\
\bottomrule
\end{tabular}
\caption{Results under GRPO training.}
\label{tab:grpo}
\end{table}

\section{Use Of AI Assistants}
Alpaca-Eval and MT-Bench use GPT for evaluation; aside from that, AI is used only for polishing and grammar checks.

\end{document}

%% file: latex/1_Intro.tex
\section{Introduction}%
\label{sec:intro}

Reinforcement Learning from Human Feedback (RLHF) \cite{askell2021general,ouyang2022training,ziegler1909fine,dong2024rlhfworkflowrewardmodeling} is a popular method for aligning large language models (LLMs) with human preferences, used in models like GPT \cite{openai2024gpt4technicalreport}, Qwen \cite{qwen2025qwen25technicalreport,yang2024qwen2technicalreport}, DeepSeek \cite{deepseekai2025deepseekv3technicalreport,deepseekai2025deepseekr1incentivizingreasoningcapability}, Gemini \cite{geminiteam2024gemini15unlockingmultimodal} and Llama \cite{grattafiori2024llama3herdmodels,touvron2023llama2openfoundation}. The framework involves three stages: supervised fine-tuning, reward model training via comparisons between preferred and dispreferred outputs (using methods like Bradley-Terry model \cite{bradley1952rank}), and reinforcement learning optimization \cite{schulman2017proximal}. 

However, RLHF generally suffers from reward hacking \cite{pmlr-v202-gao23h,weng2024rewardhack}, where policy learning leverages flaws in the trained reward model to maximize reward scores but does not learn the true human preferences. Empirical analysis reveals that reward hacking manifests through multiple mechanisms: (1) Explicit surface-level bias, such as reward model usually favoring longer responses~\cite{singhal2023long} or preferring particular response formats (\textit{e.g.}, numbered lists or markdown tables) \cite{zhang2024listsemojisformatbias}; (2) Implicit semantic bias, which arises from latent correlations in the training data distribution, where the reward model learns to associate higher reward with specific syntactic structures or topic distributions that match frequent patterns in the preference dataset ~\cite{pang2023reward,openai2025sycophancy}.

A particularly prevalent form of reward hacking is length bias~\cite{singhal2023long,NEURIPS2023_949f0f8f}, where reward models favor longer outputs over shorter ones. This bias not only distorts the reward model's preference modeling but also leads to excessively verbose generations in reinforcement learning finetuned models. A key factor of this problem lies in human preference data, which often exhibits bias and inconsistencies due to challenges such as imperfect rating criteria and variability in annotator quality~\cite{chen2024odin,NEURIPS2020_1f89885d,pang2023reward,lambert2023alignment}. Specifically, with respect to length, human raters disproportionately favor longer outputs, a tendency that reward models can exploit, thereby causing length bias. Given the inherent difficulties in obtaining perfectly reliable human annotations, developing algorithmic approaches to mitigate such spurious correlations becomes increasingly crucial.

Existing approaches sometimes do not characterize the bias form. For instance, RRM \cite{liu2025rrm} adopts a causal framework to achieve a more balanced data distribution, while other methods incorporate KL regularization terms during policy training \cite{NEURIPS2020_1f89885d,ouyang2022training}. Alternatively, other studies assume a linear length-reward relation for tractability. ODIN \cite{chen2024odin}, for example, introduces a dual-headed architecture designed to decouple length-dependent scoring from quality-based assessment, and uses the Pearson correlation coefficient \cite{pearson1895regression} to quantify the length-reward relation. Similarly, length penalty \cite{singhal2023long} directly subtracts the product of length and a constant from the reward to mitigate bias. Additionally, Huang \textit{et.al.} \cite{huang2025posthoc} calculate the hacked reward by performing linear regression on points within a certain neighborhood during the reward model inference phase. Although the linear assumption offers mathematical simplicity and intuitive feasibility, it fails to capture some details, like non-linear features where length interacts with reward in complex ways.

To overcome this problem, we introduce \ourm, an automated framework designed to model the complex non-linear relation between output length and reward scores, enabling more precise debiasing. The method begins by training a conventional reward model which inherently has length-related bias to warm up. Building upon this, a lightweight fitting model is trained to explicitly characterize how reward scores correlate with response length. By integrating these learned patterns into the reward model, the system effectively mitigates length bias without compromising its core preference modeling functionality.
Empirical results confirm that \ourm{} achieves a more balanced length-reward distribution. When deployed in downstream algorithms such as Direct Preference Optimization (DPO) \cite{NEURIPS2023_a85b405e} and Best-of-N (BoN) \cite{gui2024bonbon,sessa2025bond,dong2023raft}, the debiased model demonstrates improved performance on length-controlled win rates, reducing excessive verbosity while maintaining competitive task accuracy. Further analysis of the fitting process reveals a multi-stage bias pattern: exhibiting strong linear correlation for short responses, flattening of the relation for longer responses and in some cases a slight downward tendency for extended outputs. Our contributions can be summarized as:

\begin{itemize}

\item We propose a multi-stage framework that autonomously learns non-linear relation between response length and hacked reward and uses this relation to better mitigate the length bias. 
\item We demonstrate the effectiveness of our length debiasing approach through comprehensive validation, including length-reward distribution on preference dataset, length-controlled win rate and length distribution of responses selected by reward models.
\item We show the fitting result of the relation between length and hacked reward and identify that the length bias in the reward model is non-linear, which further illustrates the importance of debiasing with non-linear relations. 

\end{itemize}

%% file: latex/2_RelatedWorks.tex
\section{Related Work}%
\label{sec:relatedwork}
\paragraph{Reinforcement Learning From Human Feedback}RLHF \cite{askell2021general,ouyang2022training,ziegler1909fine,dong2024rlhfworkflowrewardmodeling} is an optimization algorithm proposed to align with human preferences. This algorithm is often diverse, with the most basic one being PPO \cite{schulman2017proximal}. Building upon PPO, several improved methods have been derived: GRPO \cite{shao2024deepseekmath} optimizes strategies through relative reward comparisons among multiple candidate outputs within a group, eliminating the need for a separate value model; DAPO \cite{yu2025dapoopensourcellmreinforcement} addresses issues such as entropy collapse, reward noise, and training instability in GRPO; and BoN~\cite{gui2024bonbon,sessa2025bond,dong2023raft} directly utilizes the reward model to select the one with the highest reward score as the final output. Additionally, DPO \cite{NEURIPS2023_a85b405e} unifies reward modeling and reinforcement learning into a single stage by directly optimizing the policy on pairwise preference data, eliminating the need for an explicit reward model and online sampling. Simpo \cite{NEURIPS2024_e099c1c9} further enhances model performance by removing the reference model and incorporating target reward boundaries and length normalization on the basis of DPO. Apart from them, there are many excellent studies that have contributed to the RLHF \cite{chen2024noisecontrastivealignmentlanguage,hong-etal-2024-orpo,pmlr-v238-gheshlaghi-azar24a,ethayarajh2024kto,richemond2024offlineregularisedreinforcementlearning}.

 \paragraph{Length Bias in Reward Hacking} A typical example of reward hacking is length bias, where the reward model prefers longer responses irrespective of actual response quality, leading the trained policy to generate unnecessarily verbose outputs. A part of existing approaches alleviate length bias through comprehensive reward hacking mitigation strategies, like some incorporate KL regularization terms during policy training \cite{NEURIPS2020_1f89885d,ouyang2022training}. Additionally, Eisenstein \textit{et.al.} \cite{eisenstein2024helping} point out that reward model ensembles can alleviate reward hacking and WARP \cite{ramé2024warpbenefitsweightaveraged} as well as WARM \cite{ramé2024warmbenefitsweightaveraged} utilize model merging techniques to reduce reward hacking, whereas RRM \cite{liu2025rrm} introduces a data augmentation approach by incorporating a causal framework to alleviate the hacking. Apart from these, others specifically target length debiasing, such as length penalty \cite{singhal2023long} directly subtracts the product of length and a certain coefficient from the reward to debias in a simple and intuitive way. Shen \textit{et.al.}~\cite{shen2023loose} applying Product-of-Experts to decouple the length and reward. Huang \textit{et.al.} \cite{huang2025posthoc} derive the hacked reward by applying linear regression to nearby points in the reward model's inference stage. Park \textit{et al.} \cite{park-etal-2024-disentangling} introduce a length regularization term into the DPO objective, and SamPO~\cite{lu-etal-2024-eliminating} address the cumulative length-dependent structure of KL divergence through per-token normalization in DPO.
 Moreover, ODIN \cite{chen2024odin} decoupling the reward model's scoring for length and quality with two-head structure to mitigate length bias. ALBM~\cite{bu-etal-2025-beyond} further introduces a prompt analyzer that adaptively reweights the reward based on the prompt. These works either do not explicitly model the form of length bias or directly assume a linear relation between them. Therefore in this paper we continue to focus on length debiasing but moves beyond the simplistic assumption of a linear relation between length and reward bias. Instead, we employ a dedicated lightweight model to directly fit this relation, enabling more precise length debiasing based on an accurate understanding of the bias.

%% file: latex/3_Method.tex
\section{Method}%
\label{sec:method}

\begin{figure*}[ht]
  
  \centering
  \includegraphics[width=1\textwidth]{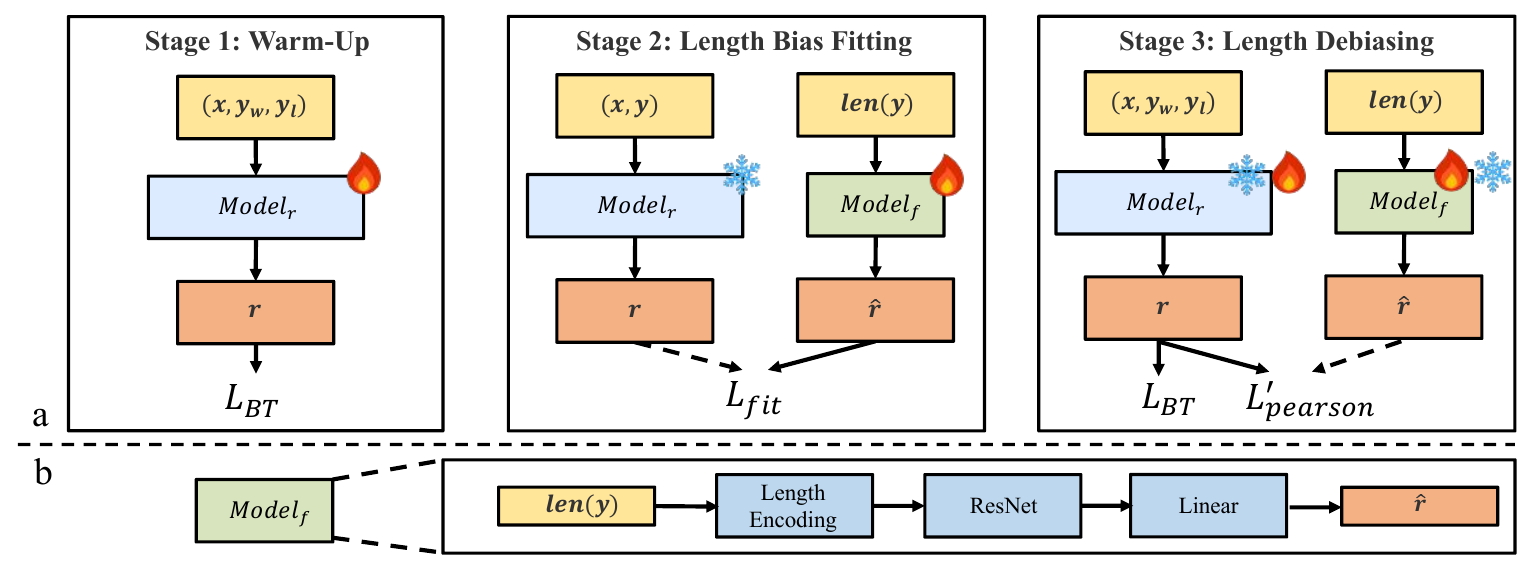}
  \caption{\textbf{a:} An overview of our method. We first use traditional reward model training to initially establish the model's length bias. Then we employ a lightweight fitting model to fit the reward hacking: given the length of a response, we minimize loss to make the output of the fitting model as close as possible to that of the reward model. The final step involves debiasing the length bias in the reward model based on the relation fitted by the fitting model, these two models take turns to train. \textbf{b:} The detailed architecture of the $model_f$.}

  \label{fig:overview}
\end{figure*}

\label{sub:Overview}

This section elaborates on our whole framework as shown in Figure~\ref{fig:overview}(a), which consists of three key stages: (1) a warm-up stage, where a standard reward model is trained and retains its inherent length bias,  
(2) a bias-fitting stage, where a lightweight model learns the relation between response length and biased reward, thereby capturing non-linear bias patterns, and  
(3) a debiasing stage, where the fitted relation is incorporated to debias.

Additionally, our framework employs two distinct models: 
(1) Reward model  ($model_{r}(x,y)$ or $r(x,y)$ for simplicity), which serves as a reward function for response quality. This model is initialized from an existing large language models. (2) Fitting model ($model_{f}(y)$), a lightweight model designed to fit the length bias inherent in the reward model $r(x, y)$.
\subsection{Warm-Up}%

The primary objective of the warm-up stage is to obtain a reward model with inherent length bias, a systematic tendency to assign a higher score to longer responses before implementing corrective measures. We initialize training using the standard reward modeling paradigm based on the Bradley-Terry model~\cite{bradley1952rank}, where given an input prompt $x$, the human-preferred response is denoted as $y_w$ and the dispreferred response as $y_l$. The training loss function is:
\begin{equation}
    \mathcal{L}_{\text{BT}} = -\mathbb{E}_{(x,y_w,y_l)}\left[\log \sigma\left(r(x,y_w) - r(x,y_l)\right)\right].
\end{equation}
Previous approaches operating under the assumption of an approximate linear relation between response length and reward, so they could directly apply length debiasing during initial training. In contrast, our method requires a precise characterization of this relation. The warm-up stage deliberately preserves length bias to enable subsequent learning and systematic removal of this bias.

\subsection{Length Bias Fitting}%

After training a reward model with inherent length bias, we proceed to formally characterize and mitigate this relation using the fitting model. The proposed approach operates as follows: given the input response length $\text{len}(y)$, we first project this scalar value into a $d$-dimensional features (in our training $d$ equals to $32$). Inspired by positional encoding (PE) \cite{NIPS2017_3f5ee243}, we treat the response length as the position in PE, thereby deriving a length encoding (LE) that embeds length information into features. Subsequently, the features after encoding are then processed through ResNet~\cite{He_2016_CVPR} architecture, the output is then fed into a final linear projection layer which serves as the regression head to produce the predicted reward $\hat{r}$ (Figure \ref{fig:overview}(b)). The optimization objective of the $model_{f}$ is to minimize the discrepancy between the predicted reward $\hat{r}$ and the actual reward output $r$. We formulate the loss function as:
\begin{equation}
    \label{eq:pearson}
\mathcal{L}_\textit{fit}= -\mathcal{L}_\textit{pearson}, \mathcal{L}_\textit{pearson}= \left|\rho\left(r_\textit{sg}, \hat{r}\right)\right|,
\end{equation}
\begin{equation}
\hat{r} = model_{f}(\text{len}(y)),~ 
r_\textit{sg} = model_{r}(x,y). 
\end{equation}
The $\rho\left(r_\textit{sg}, \hat{r}\right)$ in Equation \ref{eq:pearson} means the Pearson correlation coefficient \cite{pearson1895regression}, which could be calculated by:
\begin{equation}
\rho(r,\hat{r}) = \frac{\sum_{i=1}^n (r_i - \bar{r})(\hat{r}_i - \bar{\hat{r}})}{\sqrt{\sum_{i=1}^n (r_i - \bar{r})^2} \sqrt{\sum_{i=1}^n (\hat{r}_i - \bar{\hat{r}})^2}}.
\end{equation}
Here $r_i$ denotes the reward assigned by the reward model to the $i$-th sample within the current batch, and $\bar{r}$ is the corresponding batch mean; analogously, $\hat{r}_i$ and $\bar{\hat{r}}$ denote the $i$-th predicted reward and the batch mean. 
To prevent gradient flow through the correlation term, we introduce $r_{sg}$ as a stop-gradient copy of $r$, treating it as a constant in the computation. The Pearson loss $\mathcal{L}_\textit{pearson}$ is designed to maximize the correlation between $\hat{r}$ and $r_\textit{sg}$, thereby driving the fitting model to accurately capture the reward model’s length bias. In this formulation, the Pearson correlation coefficient is just employed as an optimization objective to capture the dependency between the two rewards, without implying assumption of linearity in length-reward relation. The fitting model itself is responsible for learning the underlying non-linear relation. In other words, the Pearson term is used only to align the correlation pattern between the two rewards, while the nonlinear fitting model independently learns the actual reward-length dependency.

\subsection{Length Debiasing}%

After fitting the length bias through the fitting model, we now debias the reward model  that was initially trained in warm-up stage by incorporating two critical objectives into its training: (1) preserving its discriminative capacity for human preferences, and (2) decoupling its outputs from response length dependence. This is achieved through a composite loss function:
\begin{equation}
\mathcal{L}_{debiased} 
 =\mathcal{L}_\textit{pearson}^{'} + \mathcal{L}_\textit{BT}.
\end{equation}
Compared to $\mathcal{L}_{\textit{pearson}}$, $\mathcal{L}_{\textit{pearson}}^{'}$ is slightly adapted to ensure gradient backpropagation through the reward model:
\begin{equation}
\mathcal{L}_\textit{pearson}^{'}=|\rho(r, \hat{r}_\textit{sg})|.
\end{equation}
The $\mathcal{L}_{\textit{pearson}}^{'}$ is to make the output of the reward model as uncorrelated as possible to the predicted reward of the fitting model, and the $\mathcal{L}_\textit{BT}$ is to ensure that the model still has the ability to model human preferences. Additionally, to better fit the bias of the model, these two models take turns to train and the loss function could be written as:
\begin{equation}
\mathcal{L}
 =I(step)*\mathcal{L}_\textit{debiased} + (1-I(step))*\mathcal{L}_\textit{fit}.
\end{equation}
Here $I(step)$ is the indicator function that indicates which model is trained under this step. For example, if we use every $a$ (in our training $a$ equals to $8$) steps to change the model for training in the third stage, then $I(step)$ can be expressed as:
\begin{equation}
\scalebox{0.9}{$
I(step)=
\left\{
\begin{aligned}
    &0,~2ka\leq step< 2ka+a,~k\in \mathbb{N},\\
    &1,~2ka+a\leq step< 2(k+1)a,~k\in \mathbb{N}.\\
\end{aligned}
\right.$}
\end{equation}

%% file: latex/4_Exp.tex
\section{Experiments}%
\label{sec:exp}

In this section, we introduce the experimental settings and validate the effectiveness of our method through three key steps. First, we train the reward model to demonstrate its accuracy under different subsets and plot the length-reward distribution. Next, we apply reward models to various alignment algorithms to verify its effectiveness. Finally, we conduct an analysis of the length distribution between different methods and show the fitted curve of $model_{f}$ at different steps in training. 
\subsection{Experimental Settings}

\paragraph{Training Data}For training data, we utilize the static split Dahoas-rm-static\footnote{\url{https://huggingface.co/datasets/Dahoas/rm-static}} from Anthropic's HH dataset \cite{bai2022traininghelpfulharmlessassistant}, partitioning it into three subsets: $30k$ samples for supervised fine-tuning~(SFT) of the base model, $30k$ samples for reward model training, and $8k$ samples reserved for downstream task validation. Moreover, the dataset also contains $5k$ samples for testing (in Table \ref{tab:acc}).
\paragraph{Training Details}
 In our experiments, we utilize the Qwen2.5-7B\footnote{\url{https://huggingface.co/Qwen/Qwen2.5-7B}} and Gemma2-9B\footnote{\url{https://huggingface.co/google/gemma-2-9b}} models~\cite{qwen2025qwen25technicalreport,gemmateam2024gemma2improvingopen}, training them with the DeepSpeed framework \cite{Rasley_Rajbhandari_Ruwase_He_2020}. For supervised fine-tuning, we employ a learning rate of $1\text{e-}5$ with a batch size of $8$ (per-device batch size $8$) and training model for $2$ epochs. Furthermore, all reward models are initialized from the same SFT model and trained use the learning rate of $2\text{e-}5$, a batch size of $16$ (per-device batch size $4$ with gradient accumulation steps $4$), and runs $1$ epoch. We conduct these experiments on same hardware configurations: all models are trained on $8\times$A100 GPUs. Additionally, inference operations leverage vLLM~\cite{kwon2023efficient}, with sampling configured as temperature $0.7$, top-$p$ $0.8$, top-$k$ $20$, and repetition penalty $1.1$.

\paragraph{Alignment Algorithm}Given the computational demands and hyperparameter sensitivity of PPO, we focus on the evaluation of the BoN (Best of N)~\cite{gui2024bonbon,sessa2025bond,dong2023raft} and DPO (Direct Preference Optimization)~\cite{NEURIPS2023_a85b405e} approaches. The BoN implementation selects highest-scoring responses from N seed-generated outputs (here we set $N=8$). In DPO, reward models cannot be directly applied, we utilize the original $8k$ preference pairs from the dataset but replace the human labels with predictions from reward models: both responses are scored, and the preference label is reassigned according to the score. All methods share same DPO training hyperparameters: learning rate $2\text{e-}6$, $1$ epoch, batch size $16$ (per-device batch size $2$ with gradient accumulation steps $8$), and $\beta=0.1$.

\paragraph{Compared Methods} We compare three methods: the vanilla reward model (Vanilla RM) and two length-debiasing approaches based on a linear assumption, Length Penalty  \cite{singhal2023long} and ODIN \cite{chen2024odin}. The Length Penalty directly subtracts the product of response length and a fixed coefficient from the reward, whereas ODIN employs a two-head structure that linearly disentangles the reward model’s assessments of length and quality, thereby mitigating length bias.

\paragraph{Evaluation} To address the well-documented length bias in LLM evaluation (observed even in models like GPT \cite{openai2024gpt4technicalreport}), we employ the length-controlled Alpaca-Eval \cite{dubois2024length} benchmark for length debiased performance assessment. The key metrics include:  
\begin{itemize}[leftmargin=*,topsep=0pt,itemsep=0pt, parsep=0pt]

\item Win Rate (WR): The winning rate of the aligned language model against fixed reference outputs~(here we take the SFT model outputs as references because all methods share the same SFT model), as scored by GPT-4. 
\item Length-Controlled Win Rate (LC-WR): The win rate after applying length debiasing to GPT-4’s evaluation. It is a more accurate metric that reflects the text generation quality.
 
\item Length of characters/tokens ($\text{L}_{char}/\text{L}_{token}$): Average response length in characters/tokens.   
\end{itemize}
Apart from Alpaca-Eval, we test the models trained with DPO
on other benchmarks: MT-Bench \cite{zheng2023judging} and IFEval \cite{IFEVAL}. MT-Bench is a benchmark designed to evaluate the capabilities of LLMs, particularly focusing on their performance in multi-turn conversations and instruction-following tasks, and IF-Eval specifically designed to evaluate the instruction-following capability of LLMs.
\subsection{Experimental Results}
\begin{table*}[!ht]
  \centering

    \renewcommand{\arraystretch}{0.9}
  \begin{tabular}{l*{7}{c}}
    \toprule
    \multirow{2}{*}{RM Acc~(\%)} & \multicolumn{3}{c}{\textbf{Qwen2.5-7B}} & & \multicolumn{3}{c}{\textbf{Gemma2-9B}} \\
    \cmidrule{2-4} \cmidrule{6-8} & All & C-longer & R-longer & & All & C-longer & R-longer \\
    \midrule
    Vanilla RM&70.14&\textbf{80.72 }&56.88 &&66.42&\textbf{80.86}&47.63 \\
    Length Penalty&\textbf{70.45}&79.76 &59.01 &&67.31&79.41&51.81 \\
    
    ODIN&70.08&78.61&59.50 &&\textbf{67.81} &76.81 &56.69 \\
    \ourm~(Ours)&69.75&73.60&\textbf{65.70}&&67.34&69.07&\textbf{66.38} \\ 
    \bottomrule
  \end{tabular}
    \caption{Accuracy on Preference Datasets. Our approach achieves better length balance, with the reward model showing a closer accuracy across both subsets. Notably, the C-longer subset constitutes a larger proportion (58\%) of the total dataset. Given this data distribution, prioritizing accuracy optimization for the C-longer subset may result in a misleadingly favorable assessment of overall performance. }
        \label{tab:acc}
  \end{table*}
  
\begin{figure*}[ht]
  
  \centering
  \includegraphics[width=0.95\textwidth]{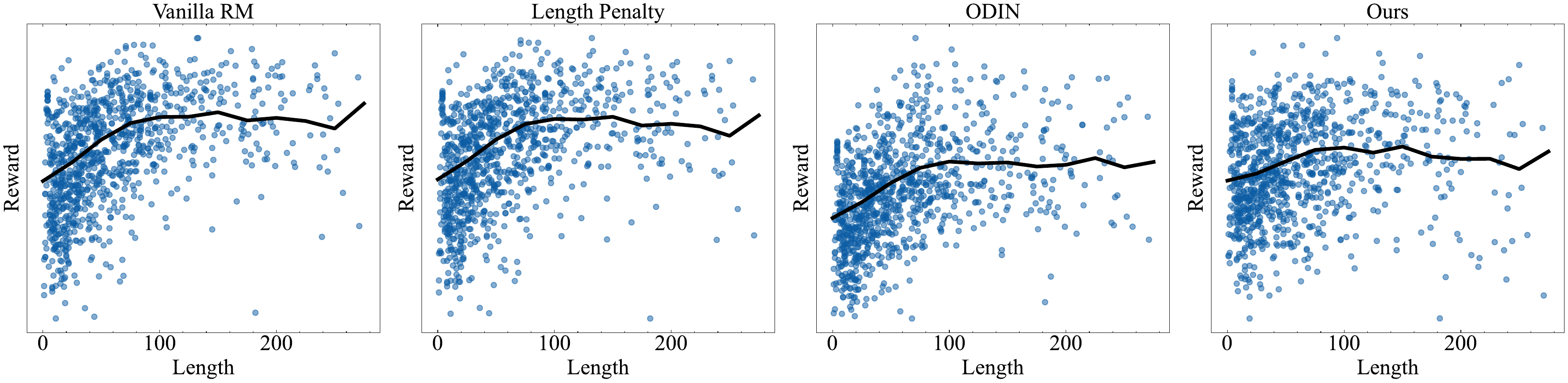}
  \caption{Scatter plot of reward-length, with binned averages 
 (black lines). Our method demonstrates a more balanced reward distribution compared to others, indicating effective debiasing. Because the absolute reward magnitude is unconstrained during RM training, it is not comparable across different reward models; therefore, we do not label the y-axis in the figure.}

  \label{fig:scatter}
\end{figure*}
  
\paragraph{Accuracy on Preference Datasets}We evaluate the accuracy of our method in preference datasets by splitting the test set into two subsets: C-longer containing samples where the chosen response was longer than the rejected response ($len(y_w) > len(y_l)$) and R-longer is the rejected response was longer ($len(y_l) > len(y_w)$). The results in Table~\ref{tab:acc} demonstrate that our approach achieves better length balance, with the reward model showing a closer accuracy across two subsets.  

As shown in Table \ref{tab:acc}, FiMi-RM reduces accuracy on the C-longer subset while improving it on R-longer. We clarify that this trade-off is not a degradation of discriminative ability. A reward model with length bias trivially achieves high C-longer accuracy by exploiting the length shortcut. In the extreme case, ranking purely by length would yield 100\% C-longer accuracy with no semantic understanding. Furthermore, since the C-longer subset constitutes a larger portion (58\%) of the total dataset, a length-biased model can inflate its overall accuracy simply by optimizing on this majority subset. So the drop in C-longer accuracy therefore indicates that FiMi-RM has successfully discarded this shortcut, and the slight reduction in overall accuracy reflects a more balanced performance distribution rather than diminished capability. Importantly, the consistent improvements across downstream benchmarks confirm that semantic quality is preserved and improved after debiasing, demonstrating that the trade-off in static preference accuracy does not hinder the performance.

\paragraph{Length-Reward Distribution}  To further analyze the test set results, we plot a scatter graph of response length \footnote{Unless otherwise specified, all subsequent references to "length" in this paper refer to token length} and reward given by different 7B models, along with the average reward for different length ranges. Since directly calculating the mean reward for each individual length results in high variance, we split the length range into bins of size $25$ to compute the average reward within each bin. As shown in Figure \ref{fig:scatter}, our method exhibits a more balanced distribution compared to other methods: Our scatter plot demonstrates better symmetry along the x-axis (length axis), and average curve is also more flatter. This figure further validate the effectiveness of our length debiasing approach.

\begin{table*}[!ht]

  \centering

  \renewcommand{\arraystretch}{0.9}
  \begin{tabular}{l *{9}{c}}
    \toprule
       \multirow{2}{*}{BoN} &\multicolumn{4}{c}{\textbf{Qwen2.5-7B}}& &\multicolumn{4}{c}{\textbf{Gemma2-9B}}  \\
      \cmidrule{2-5}  \cmidrule{7-10}
    &LC-WR$\uparrow$&WR$\uparrow$&$\text{L}_{char}$  &$\text{L}_{token}$ &       &LC-WR$\uparrow$&WR$\uparrow$&$\text{L}_{char}$  &$\text{L}_{token}$   
 \\
    \midrule
  
    Vanilla RM&68.25&73.84 &638  &148&&62.91&65.77&756&179
    \\ 
     Length Penalty&69.22&74.54&620&144 &&62.69 &65.62 &712&158
       \\
ODIN&71.38&75.58&512&115 &&63.46 &65.24 &684&168
       \\
      \ourm~(Ours) &\textbf{72.59}&\textbf{76.39} &543 &125&&\textbf{66.68}&\textbf{67.77}&534&123
   \\ 
    \bottomrule
     
  \end{tabular}
    \caption{The length-controlled Alpaca-Eval results under the BoN algorithm. Our method achieved the highest win rate (WR) and length-controlled win rate (LC-WR). While our output length in Qwen2.5-7B is slightly longer than ODIN’s, we maintain better performance in LC-WR and WR, indicating more effective bias mitigation.}
        
        \label{tab:bon}
\end{table*}
\begin{table*}[!ht]

  \centering

  \renewcommand{\arraystretch}{0.9}
  \begin{tabular}{l *{9}{c}}
    \toprule
       \multirow{2}{*}{DPO} &\multicolumn{4}{c}{\textbf{Qwen2.5-7B}}& &\multicolumn{4}{c}{\textbf{Gemma2-9B}}  \\
      \cmidrule{2-5}  \cmidrule{7-10}
    &LC-WR$\uparrow$&WR$\uparrow$&$\text{L}_{char}$  &$\text{L}_{token}$ &       &LC-WR$\uparrow$&WR$\uparrow$&$\text{L}_{char}$  &$\text{L}_{token}$   
 \\
    \midrule
  
    Vanilla RM&58.56&61.73&1089&254&&57.35&59.81&1135&267
    \\ 
  
    Length Penalty&57.67&62.89&980&227&&57.78&60.03&1077&250
    \\      
ODIN&58.91&63.16&1044& 244&&57.96 &61.19 &905&214
       \\
      \ourm~(Ours)&\textbf{62.17}&\textbf{67.32} &757&174&&\textbf{59.47}&\textbf{61.61}&773&189
   \\ 
    \bottomrule
     
  \end{tabular}
            \caption{The length-controlled Alpaca-Eval results under the DPO algorithm. Our results demonstrate better performance in both LC-WR and WR.}

        \label{tab:dpo}
  \end{table*}

\begin{figure*}[ht!]
  
  \centering
  \includegraphics[width=1\textwidth]{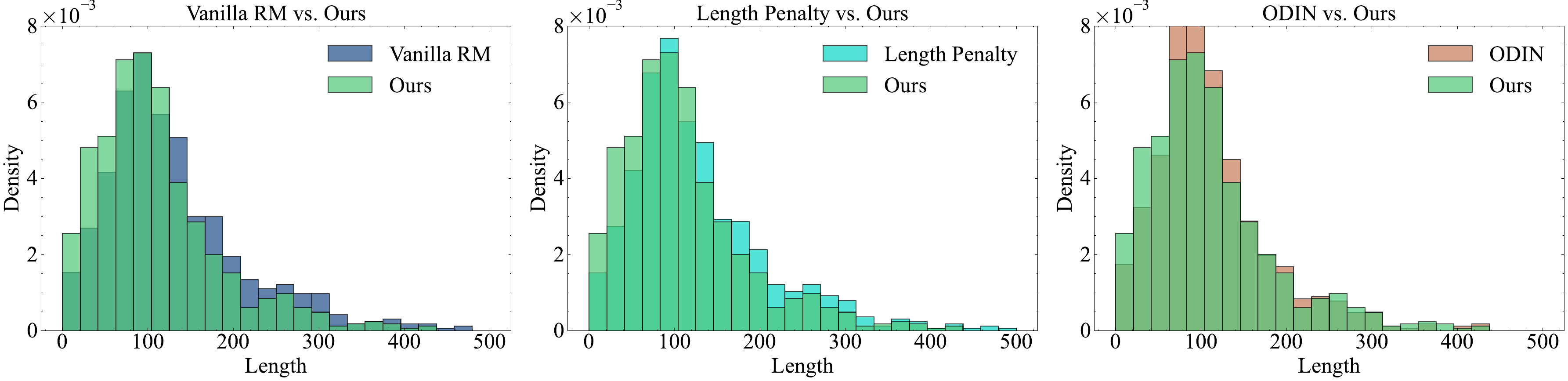}
  \caption{The pairwise comparison of the distribution of responses selected by BoN. The figure indicates that, relative to vanilla RM and Length Penalty, our approach demonstrates a stronger inclination toward shorter responses in BoN selection. Although ODIN also mitigates bias toward overly lengthy outputs, it mainly shifts preferences toward medium-length responses rather than enhancing the selection of shorter ones.  }

  \label{fig:bon}
\end{figure*}
\begin{figure*}[ht!]
  
  \centering
  \includegraphics[width=1\textwidth]{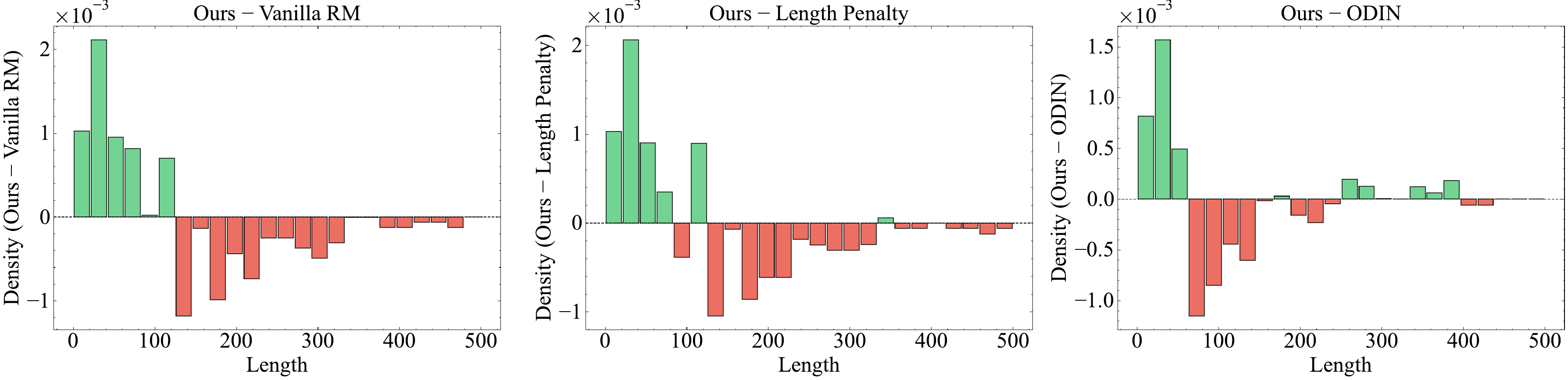}
  \caption{Difference in response length distributions under BoN, computed from the two overlapping histograms shown in Figure \ref{fig:bon}. Positive values (green) indicate that FiMi-RM assigns higher density to the corresponding length range compared to the others; negative values (red) indicate lower density.   }

  \label{fig:bon_div}
\end{figure*}

\paragraph{Performance of Different Alignment Algorithms Using Reward Models}

As shown in Table \ref{tab:bon} and  Table \ref{tab:dpo}, we conduct an evaluation under two alignment algorithms: DPO and BoN. Our results reveal that the reward model consistently achieves higher LC-WR compared to baseline methods, demonstrating its effectiveness in mitigating length bias. While non-debiased win rate also favors ours.

In terms of response length, our method exhibit shorter outputs compare to others in most case, though our BoN-generated responses are longer than ODIN’s in Qwen2.5-7B. However, shorter length does not always indicate better performance, the better length-controlled win rate confirms that our model produces optimally balanced responses within a reasonable length range.  

To further evaluate performance, we assess the language model trained with DPO on additional benchmarks (Table \ref{tab:other}). On MT-bench, our model outperforms competing methods in the first round (T1), second round (T2), and overall average score (Avg). It also achieves higher prompt-level accuracy on IFEval, offering further evidence that our length debiasing improves output quality.

\begin{table*}[!ht]

  \centering

    \renewcommand{\arraystretch}{0.9}

  \begin{tabular}{lcccccccc}
    \toprule
      \multirow{3}{*}{DPO} 
        & \multicolumn{4}{c}{\textbf{Qwen2.5-7B}} 
        & \multicolumn{4}{c}{\textbf{Gemma2-9B}} \\
      \cmidrule(lr){2-5} \cmidrule(lr){6-9}
        & \multicolumn{3}{c}{MT-bench} & IFEval 
        & \multicolumn{3}{c}{MT-bench} & IFEval \\
      \cmidrule(lr){2-4} \cmidrule(lr){5-5} \cmidrule(lr){6-8} \cmidrule(lr){9-9}
        & T1$\uparrow$ & T2$\uparrow$ & Avg$\uparrow$ &  $\text{Acc}_{prompt}\uparrow$
        & T1$\uparrow$ & T2$\uparrow$ & Avg$\uparrow$ & $\text{Acc}_{prompt}\uparrow$  \\
    \midrule
      Vanilla RM & 5.45 & 3.43 & 4.44 & 17.4 
           & 3.99 & 3.06 & 3.53 & 18.0 \\
     Length Penalty   & 5.11 & 3.35 & 4.23 & 16.6 
           & 4.51 & 3.18 & 3.84 & 18.2 \\
      ODIN & 5.33 & 3.73 & 4.53 & 18.6 
           & 4.46 & 3.16 & 3.81 & 17.8 \\
      \ourm~(Ours) & \textbf{5.60} & \textbf{4.04} & \textbf{4.82} & \textbf{19.0}
           & \textbf{4.65} & \textbf{3.34} & \textbf{3.99} & \textbf{19.6} \\
    \bottomrule
  \end{tabular}
    \caption{Other benchmarks' results under the DPO algorithm. Our model outperforms other methods on MT-bench in both the first round (T1), the second round (T2), and the average score. It also achieves improvements on prompt-level accuracy in IFEval, further demonstrating that our method enhances output quality after length debiasing.}
  \label{tab:other}
\end{table*}

\begin{figure*}[ht]
  
  \centering
  \includegraphics[width=1\textwidth]{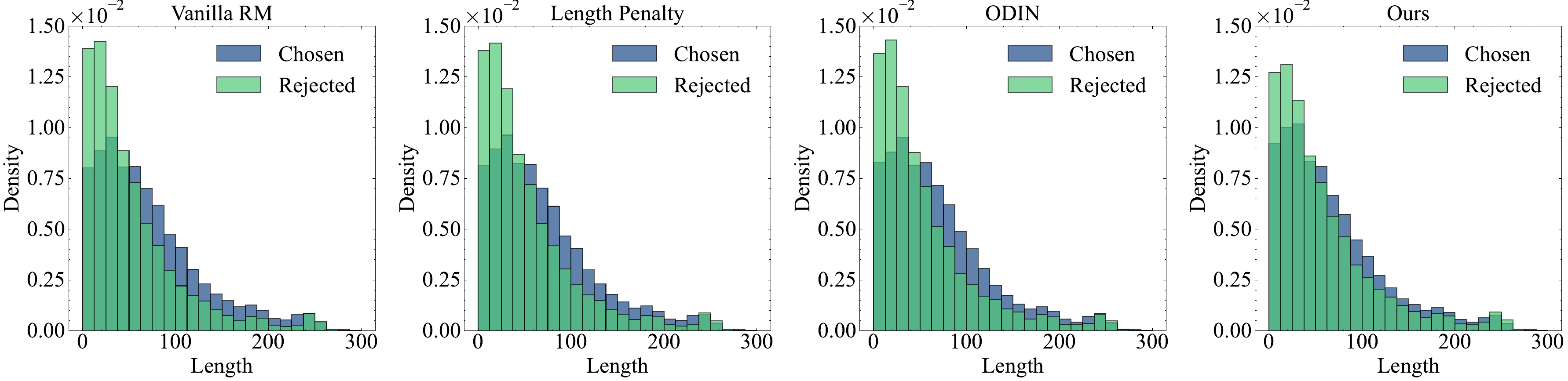}
  \caption{Length distribution of chosen and rejected responses in the labeling stage of DPO. The gap between chosen and rejected response length is noticeably smaller for our method when comparing to others. }

  \label{fig:dpo}
\end{figure*}

\begin{figure*}[ht]

  \centering
  \includegraphics[width=1\textwidth]{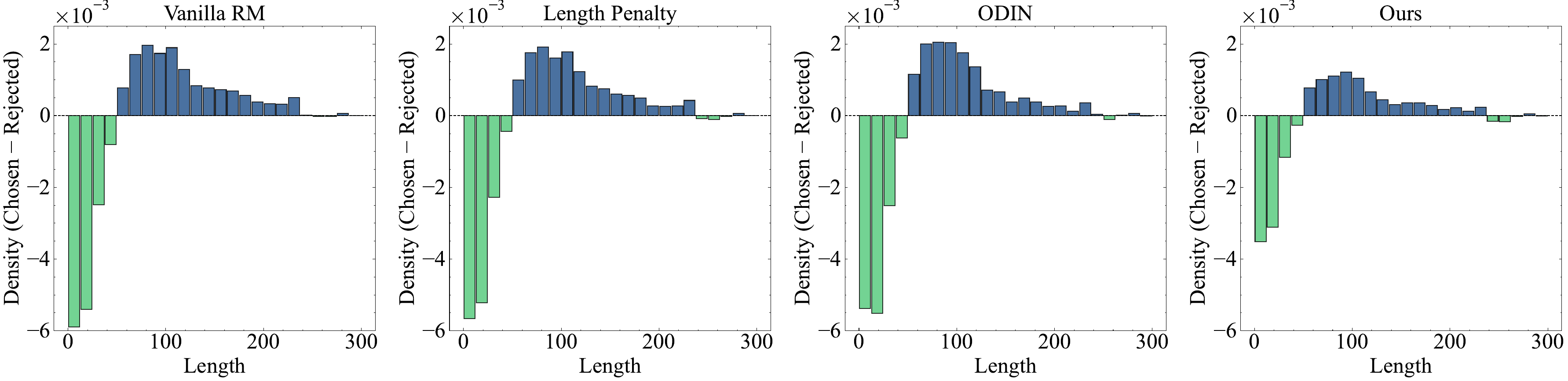}
  \caption{The difference between the chosen and rejected response length distributions (chosen - rejected) shown in~Figure \ref{fig:dpo}. }

  \label{fig:dpo_div}
\end{figure*}

\begin{figure*}[ht!]
  
  \centering
  \includegraphics[width=\textwidth]{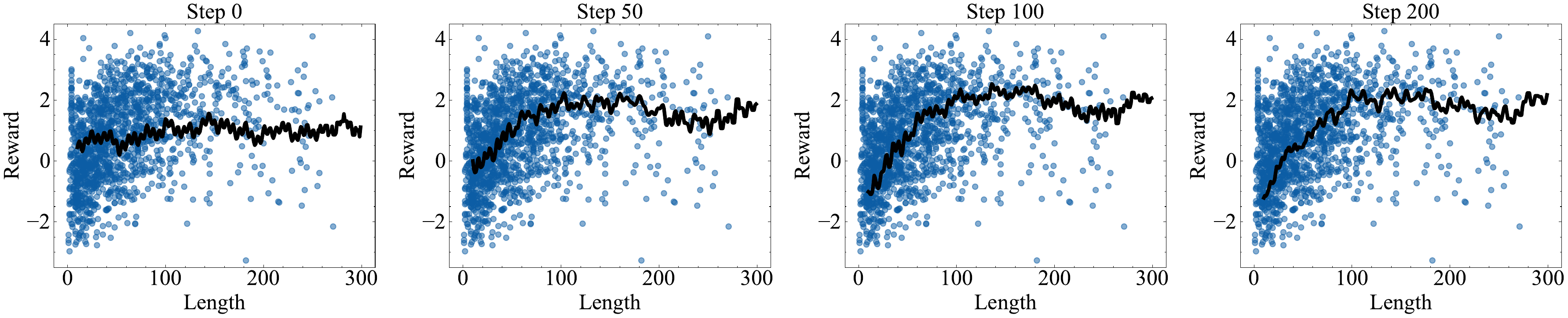}
  \caption{The fitted curve of $model_{f}$ at different steps in training.
}

  \label{fig:relation}
\end{figure*}
\paragraph{Length Distribution of Reward Models' Selection}  

We analyze the length distribution of responses selected by different 7B reward models under the BoN algorithm, which is shown in Figure \ref{fig:bon} and Figure \ref{fig:bon_div}, as well as the distribution of chosen ($y_w$) and rejected ($y_l$) responses during DPO data annotation, which is shown in Figure \ref{fig:dpo} and \ref{fig:dpo_div}.  

The results in Figure \ref{fig:bon} and \ref{fig:bon_div} show that, compared to vanilla RM and Length Penalty, our method exhibits stronger preferences for shorter responses in BoN selection. While ODIN also reduces bias toward excessively long outputs, it primarily shifts preferences toward medium-length responses rather than increasing selection of shorter ones. In contrast, our approach has a more balanced distribution, with a  tendency to favor concise outputs.

Furthermore, in DPO annotation (Figure \ref{fig:dpo} and \ref{fig:dpo_div}), the gap between chosen and rejected response length distribution is smaller for our method compared to other methods. This indicates that our reward model has less length bias in preference data labeling. The reduced discrepancy between chosen and rejected length further validates that our strategy mitigates length reward hacking better.

\paragraph{Training Process of Fitting Model}

In Figure \ref{fig:relation} we show the fitted curve of $model_{f}$ at different steps in training stage 2 (stage of length bias fitting). The blue scatter points represent the actual output of the reward model before debiasing, while the black curves illustrate the relation fitted by the fitting model. Here we show the process of the fitting model capturing the bias relation and the fitted result. As shown in the figure, the initially fitted curve (step 0) exhibits no clear pattern at the beginning because it is randomly initialized, and with training progresses, the curve gradually aligns with the trend of the scatter points.

Furthermore, the last subfigure (step 200) reveals two distinct phases: For responses shorter than 100 tokens, the reward-length relation exhibits a strong linear trend, indicating that the bias grows almost proportionally with length. When beyond 100 tokens, the relation becomes noticeably flatter and in some cases even shows a slight downward tendency, suggesting that the effect of length on reward diminishes or reverses for longer outputs.

%% file: latex/5_conclusion.tex
\section{Conclusion}%
\label{sec:conclusion}
This paper primarily investigates length debiasing in reward models. Previous approaches to length debiasing are typically not characterize the bias form or assume a linear relation between input length and the hacked reward. To achieve better length debiasing, we employ a lightweight model to explicitly fit the relation between input length and the hacked reward from the reward model. In experiments, we test our method on BoN and DPO, observing improvements in different benchmarks. From the perspective of societal impact, better aligning models with human preferences helps them serve humanity more effectively and safely. However, stronger alignment capabilities could also be misused to align with harmful content. Therefore, we must strengthen the regulatory.

%% file: custom.bib
@misc{ziegler1909fine,
      title={Fine-Tuning Language Models from Human Preferences}, 
      author={Daniel M. Ziegler and Nisan Stiennon and Jeffrey Wu and Tom B. Brown and Alec Radford and Dario Amodei and Paul Christiano and Geoffrey Irving},
      year={2020},
      eprint={1909.08593},
      archivePrefix={arXiv},
      primaryClass={cs.CL},
      url={https://arxiv.org/abs/1909.08593}, 
}

@article{ouyang2022training,
  title={Training language models to follow instructions with human feedback},
  author={Long Ouyang and Jeff Wu and Xu Jiang and Diogo Almeida and Carroll L. Wainwright and Pamela Mishkin and Chong Zhang and Sandhini Agarwal and Katarina Slama and Alex Ray and John Schulman and Jacob Hilton and Fraser Kelton and Luke Miller and Maddie Simens and Amanda Askell and Peter Welinder and Paul Christiano and Jan Leike and Ryan Lowe},
  journal={Advances in neural information processing systems},
  volume={35},
  pages={27730--27744},
  year={2022}
}

@article{askell2021general,
  title={A general language assistant as a laboratory for alignment},
  author={Amanda Askell and Yuntao Bai and Anna Chen and Dawn Drain and Deep Ganguli and Tom Henighan and Andy Jones and Nicholas Joseph and Ben Mann and Nova DasSarma and Nelson Elhage and Zac Hatfield-Dodds and Danny Hernandez and Jackson Kernion and Kamal Ndousse and Catherine Olsson and Dario Amodei and Tom Brown and Jack Clark and Sam McCandlish and Chris Olah and Jared Kaplan},
  journal={arXiv preprint arXiv:2112.00861},
  year={2021}
}

@article{bradley1952rank,
  title={Rank analysis of incomplete block designs: I. The method of paired comparisons},
  author={Bradley, Ralph Allan and Terry, Milton E},
  journal={Biometrika},
  volume={39},
  number={3/4},
  pages={324--345},
  year={1952},
  publisher={JSTOR}
}

@article{singhal2023long,
  title={A long way to go: Investigating length correlations in rlhf},
  author={Singhal, Prasann and Goyal, Tanya and Xu, Jiacheng and Durrett, Greg},
  journal={arXiv preprint arXiv:2310.03716},
  year={2023}
}

@inproceedings{
chen2024odin,
title={{ODIN}: Disentangled Reward Mitigates Hacking in {RLHF}},
author={Lichang Chen and Chen Zhu and Jiuhai Chen and Davit Soselia and Tianyi Zhou and Tom Goldstein and Heng Huang and Mohammad Shoeybi and Bryan Catanzaro},
booktitle={Forty-first International Conference on Machine Learning},
year={2024},
url={https://openreview.net/forum?id=zcIV8OQFVF}
}

@article{dubois2024length,
  title={Length-Controlled AlpacaEval: A Simple Way to Debias Automatic Evaluators},
  author={Dubois, Yann and Galambosi, Bal{\'a}zs and Liang, Percy and Hashimoto, Tatsunori B},
  journal={arXiv preprint arXiv:2404.04475},
  year={2024}
}

@InProceedings{pmlr-v202-gao23h,
  title = 	 {Scaling Laws for Reward Model Overoptimization},
  author =       {Gao, Leo and Schulman, John and Hilton, Jacob},
  booktitle = 	 {Proceedings of the 40th International Conference on Machine Learning},
  pages = 	 {10835--10866},
  year = 	 {2023},
  editor = 	 {Krause, Andreas and Brunskill, Emma and Cho, Kyunghyun and Engelhardt, Barbara and Sabato, Sivan and Scarlett, Jonathan},
  volume = 	 {202},
  series = 	 {Proceedings of Machine Learning Research},
  month = 	 {23--29 Jul},
  publisher =    {PMLR},
  url = 	 {https://proceedings.mlr.press/v202/gao23h.html},
  
}

@article{schulman2017proximal,
  title={Proximal policy optimization algorithms},
  author={Schulman, John and Wolski, Filip and Dhariwal, Prafulla and Radford, Alec and Klimov, Oleg},
  journal={arXiv preprint arXiv:1707.06347},
  year={2017}
}

@article{shao2024deepseekmath,
  title={Deepseekmath: Pushing the limits of mathematical reasoning in open language models},
  author={Zhihong Shao and Peiyi Wang and Qihao Zhu and Runxin Xu and Junxiao Song and Xiao Bi and Haowei Zhang and Mingchuan Zhang and Y. K. Li and Y. Wu and Daya Guo},
  journal={arXiv preprint arXiv:2402.03300},
  year={2024}
}

@misc{yu2025dapoopensourcellmreinforcement,
      title={DAPO: An Open-Source LLM Reinforcement Learning System at Scale}, 
      author={Qiying Yu and Zheng Zhang and Ruofei Zhu and Yufeng Yuan and Xiaochen Zuo and Yu Yue and Weinan Dai and Tiantian Fan and Gaohong Liu and Lingjun Liu and Xin Liu and Haibin Lin and Zhiqi Lin and Bole Ma and Guangming Sheng and Yuxuan Tong and Chi Zhang and Mofan Zhang and Wang Zhang and Hang Zhu and Jinhua Zhu and Jiaze Chen and Jiangjie Chen and Chengyi Wang and Hongli Yu and Yuxuan Song and Xiangpeng Wei and Hao Zhou and Jingjing Liu and Wei-Ying Ma and Ya-Qin Zhang and Lin Yan and Mu Qiao and Yonghui Wu and Mingxuan Wang},
      year={2025},
      eprint={2503.14476},
      archivePrefix={arXiv},
      primaryClass={cs.LG},
      url={https://arxiv.org/abs/2503.14476}, 
}

@article{dong2023raft,
  title={Raft: Reward ranked finetuning for generative foundation model alignment},
  author={Dong, Hanze and Xiong, Wei and Goyal, Deepanshu and Zhang, Yihan and Chow, Winnie and Pan, Rui and Diao, Shizhe and Zhang, Jipeng and Shum, Kashun and Zhang, Tong},
  journal={arXiv preprint arXiv:2304.06767},
  year={2023}
}

@inproceedings{
sessa2025bond,
title={{BOND}: Aligning {LLM}s with Best-of-N Distillation},
author={Pier Giuseppe Sessa and Robert Dadashi-Tazehozi and Leonard Hussenot and Johan Ferret and Nino Vieillard and Alexandre Rame and Bobak Shahriari and Sarah Perrin and Abram L. Friesen and Geoffrey Cideron and Sertan Girgin and Piotr Stanczyk and Andrea Michi and Danila Sinopalnikov and Sabela Ramos Garea and Am{\'e}lie H{\'e}liou and Aliaksei Severyn and Matthew Hoffman and Nikola Momchev and Olivier Bachem},
booktitle={The Thirteenth International Conference on Learning Representations},
year={2025},
url={https://openreview.net/forum?id=0tAXMiSufG}
}

@article{gui2024bonbon,
  title={Bonbon alignment for large language models and the sweetness of best-of-n sampling},
  author={Gui, Lin and G{\^a}rbacea, Cristina and Veitch, Victor},
  journal={arXiv preprint arXiv:2406.00832},
  year={2024}
}

@inproceedings{NEURIPS2023_a85b405e,
 author = {Rafailov, Rafael and Sharma, Archit and Mitchell, Eric and Manning, Christopher D and Ermon, Stefano and Finn, Chelsea},
 booktitle = {Advances in Neural Information Processing Systems},
 editor = {A. Oh and T. Naumann and A. Globerson and K. Saenko and M. Hardt and S. Levine},
 pages = {53728--53741},
 publisher = {Curran Associates, Inc.},
 title = {Direct Preference Optimization: Your Language Model is Secretly a Reward Model},
 url = {https://proceedings.neurips.cc/paper_files/paper/2023/file/a85b405ed65c6477a4fe8302b5e06ce7-Paper-Conference.pdf},
 volume = {36},
 year = {2023}
}

@inproceedings{NEURIPS2024_e099c1c9,
 author = {Meng, Yu and Xia, Mengzhou and Chen, Danqi},
 booktitle = {Advances in Neural Information Processing Systems},
 editor = {A. Globerson and L. Mackey and D. Belgrave and A. Fan and U. Paquet and J. Tomczak and C. Zhang},
 pages = {124198--124235},
 publisher = {Curran Associates, Inc.},
 title = {SimPO: Simple Preference Optimization with a Reference-Free Reward},
 url = {https://proceedings.neurips.cc/paper_files/paper/2024/file/e099c1c9699814af0be873a175361713-Paper-Conference.pdf},
 volume = {37},
 year = {2024}
}

@misc{ramé2024warpbenefitsweightaveraged,
      title={WARP: On the Benefits of Weight Averaged Rewarded Policies}, 
      author={Alexandre Ramé and Johan Ferret and Nino Vieillard and Robert Dadashi and Léonard Hussenot and Pierre-Louis Cedoz and Pier Giuseppe Sessa and Sertan Girgin and Arthur Douillard and Olivier Bachem},
      year={2024},
      eprint={2406.16768},
      archivePrefix={arXiv},
      primaryClass={cs.LG},
      url={https://arxiv.org/abs/2406.16768}, 
}

@misc{ramé2024warmbenefitsweightaveraged,
      title={WARM: On the Benefits of Weight Averaged Reward Models}, 
      author={Alexandre Ramé and Nino Vieillard and Léonard Hussenot and Robert Dadashi and Geoffrey Cideron and Olivier Bachem and Johan Ferret},
      year={2024},
      eprint={2401.12187},
      archivePrefix={arXiv},
      primaryClass={cs.LG},
      url={https://arxiv.org/abs/2401.12187}, 
}

@inproceedings{
liu2025rrm,
title={{RRM}:  Robust Reward Model Training Mitigates Reward Hacking},
author={Tianqi Liu and Wei Xiong and Jie Ren and Lichang Chen and Junru Wu and Rishabh Joshi and Yang Gao and Jiaming Shen and Zhen Qin and Tianhe Yu and Daniel Sohn and Anastasia Makarova and Jeremiah Zhe Liu and Yuan Liu and Bilal Piot and Abe Ittycheriah and Aviral Kumar and Mohammad Saleh},
booktitle={The Thirteenth International Conference on Learning Representations},
year={2025},
url={https://openreview.net/forum?id=88AS5MQnmC}
}

@InProceedings{He_2016_CVPR,
author = {He, Kaiming and Zhang, Xiangyu and Ren, Shaoqing and Sun, Jian},
title = {Deep Residual Learning for Image Recognition},
booktitle = {Proceedings of the IEEE Conference on Computer Vision and Pattern Recognition},
month = {June},
year = {2016}
}

@inproceedings{NIPS2017_3f5ee243,
 author = {Vaswani, Ashish and Shazeer, Noam and Parmar, Niki and Uszkoreit, Jakob and Jones, Llion and Gomez, Aidan N and Kaiser, \L ukasz and Polosukhin, Illia},
 booktitle = {Advances in Neural Information Processing Systems},
 editor = {I. Guyon and U. Von Luxburg and S. Bengio and H. Wallach and R. Fergus and S. Vishwanathan and R. Garnett},
 pages = {},
 publisher = {Curran Associates, Inc.},
 title = {Attention is All you Need},
 url = {https://proceedings.neurips.cc/paper_files/paper/2017/file/3f5ee243547dee91fbd053c1c4a845aa-Paper.pdf},
 volume = {30},
 year = {2017}
}

@misc{bai2022traininghelpfulharmlessassistant,
      title={Training a Helpful and Harmless Assistant with Reinforcement Learning from Human Feedback}, 
      author={Yuntao Bai and Andy Jones and Kamal Ndousse and Amanda Askell and Anna Chen and Nova DasSarma and Dawn Drain and Stanislav Fort and Deep Ganguli and Tom Henighan and Nicholas Joseph and Saurav Kadavath and Jackson Kernion and Tom Conerly and Sheer El-Showk and Nelson Elhage and Zac Hatfield-Dodds and Danny Hernandez and Tristan Hume and Scott Johnston and Shauna Kravec and Liane Lovitt and Neel Nanda and Catherine Olsson and Dario Amodei and Tom Brown and Jack Clark and Sam McCandlish and Chris Olah and Ben Mann and Jared Kaplan},
      year={2022},
      eprint={2204.05862},
      archivePrefix={arXiv},
      primaryClass={cs.CL},
      url={https://arxiv.org/abs/2204.05862}, 
}

@misc{yang2024qwen2technicalreport,
      title={Qwen2 Technical Report}, 
      author={An Yang and Baosong Yang and Binyuan Hui and Bo Zheng and Bowen Yu and Chang Zhou and Chengpeng Li and Chengyuan Li and Dayiheng Liu and Fei Huang and Guanting Dong and Haoran Wei and Huan Lin and Jialong Tang and Jialin Wang and Jian Yang and Jianhong Tu and Jianwei Zhang and Jianxin Ma and Jianxin Yang and Jin Xu and Jingren Zhou and Jinze Bai and Jinzheng He and Junyang Lin and Kai Dang and Keming Lu and Keqin Chen and Kexin Yang and Mei Li and Mingfeng Xue and Na Ni and Pei Zhang and Peng Wang and Ru Peng and Rui Men and Ruize Gao and Runji Lin and Shijie Wang and Shuai Bai and Sinan Tan and Tianhang Zhu and Tianhao Li and Tianyu Liu and Wenbin Ge and Xiaodong Deng and Xiaohuan Zhou and Xingzhang Ren and Xinyu Zhang and Xipin Wei and Xuancheng Ren and Xuejing Liu and Yang Fan and Yang Yao and Yichang Zhang and Yu Wan and Yunfei Chu and Yuqiong Liu and Zeyu Cui and Zhenru Zhang and Zhifang Guo and Zhihao Fan},
      year={2024},
      eprint={2407.10671},
      archivePrefix={arXiv},
      primaryClass={cs.CL},
      url={https://arxiv.org/abs/2407.10671}, 
}

@misc{qwen2025qwen25technicalreport,
      title={Qwen2.5 Technical Report}, 
      author={Qwen and An Yang and Baosong Yang and Beichen Zhang and Binyuan Hui and Bo Zheng and Bowen Yu and Chengyuan Li and Dayiheng Liu and Fei Huang and Haoran Wei and Huan Lin and Jian Yang and Jianhong Tu and Jianwei Zhang and Jianxin Yang and Jiaxi Yang and Jingren Zhou and Junyang Lin and Kai Dang and Keming Lu and Keqin Bao and Kexin Yang and Le Yu and Mei Li and Mingfeng Xue and Pei Zhang and Qin Zhu and Rui Men and Runji Lin and Tianhao Li and Tianyi Tang and Tingyu Xia and Xingzhang Ren and Xuancheng Ren and Yang Fan and Yang Su and Yichang Zhang and Yu Wan and Yuqiong Liu and Zeyu Cui and Zhenru Zhang and Zihan Qiu},
      year={2025},
      eprint={2412.15115},
      archivePrefix={arXiv},
      primaryClass={cs.CL},
      url={https://arxiv.org/abs/2412.15115}, 
}

@misc{openai2025sycophancy,
  author = {OpenAI},
  title = {Expanding on what we missed with sycophancy},
  url = {https://openai.com/index/expanding-on-sycophancy/},
  urldate = {2025-05-02},
  year = {2025}
}

@misc{zhang2024listsemojisformatbias,
      title={From Lists to Emojis: How Format Bias Affects Model Alignment}, 
      author={Xuanchang Zhang and Wei Xiong and Lichang Chen and Tianyi Zhou and Heng Huang and Tong Zhang},
      year={2024},
      eprint={2409.11704},
      archivePrefix={arXiv},
      primaryClass={cs.CL},
      url={https://arxiv.org/abs/2409.11704}, 
}

@misc{weng2024rewardhack,
  title   = {Reward Hacking in Reinforcement Learning},
  author  = {Weng, Lilian},
  journal = {lilianweng.github.io},
  year    = {2024},
  month   = {Nov},
  url     = {https://lilianweng.github.io/posts/2024-11-28-reward-hacking/}
}

@inproceedings{kwon2023efficient,
  title={Efficient Memory Management for Large Language Model Serving with PagedAttention},
  author={Woosuk Kwon and Zhuohan Li and Siyuan Zhuang and Ying Sheng and Lianmin Zheng and Cody Hao Yu and Joseph E. Gonzalez and Hao Zhang and Ion Stoica},
  booktitle={Proceedings of the ACM SIGOPS 29th Symposium on Operating Systems Principles},
  year={2023}
}

@misc{openai2024gpt4technicalreport,
      title={GPT-4 Technical Report}, 
      author={OpenAI and Josh Achiam and Steven Adler and Sandhini Agarwal and Lama Ahmad and Ilge Akkaya and Florencia Leoni Aleman and Diogo Almeida and Janko Altenschmidt and Sam Altman and Shyamal Anadkat and Red Avila and Igor Babuschkin and Suchir Balaji and Valerie Balcom and Paul Baltescu and Haiming Bao and Mohammad Bavarian and Jeff Belgum and Irwan Bello and Jake Berdine and Gabriel Bernadett-Shapiro and Christopher Berner and Lenny Bogdonoff and Oleg Boiko and Madelaine Boyd and Anna-Luisa Brakman and Greg Brockman and Tim Brooks and Miles Brundage and Kevin Button and Trevor Cai and Rosie Campbell and Andrew Cann and Brittany Carey and Chelsea Carlson and Rory Carmichael and Brooke Chan and Che Chang and Fotis Chantzis and Derek Chen and Sully Chen and Ruby Chen and Jason Chen and Mark Chen and Ben Chess and Chester Cho and Casey Chu and Hyung Won Chung and Dave Cummings and Jeremiah Currier and Yunxing Dai and Cory Decareaux and Thomas Degry and Noah Deutsch and Damien Deville and Arka Dhar and David Dohan and Steve Dowling and Sheila Dunning and Adrien Ecoffet and Atty Eleti and Tyna Eloundou and David Farhi and Liam Fedus and Niko Felix and Simón Posada Fishman and Juston Forte and Isabella Fulford and Leo Gao and Elie Georges and Christian Gibson and Vik Goel and Tarun Gogineni and Gabriel Goh and Rapha Gontijo-Lopes and Jonathan Gordon and Morgan Grafstein and Scott Gray and Ryan Greene and Joshua Gross and Shixiang Shane Gu and Yufei Guo and Chris Hallacy and Jesse Han and Jeff Harris and Yuchen He and Mike Heaton and Johannes Heidecke and Chris Hesse and Alan Hickey and Wade Hickey and Peter Hoeschele and Brandon Houghton and Kenny Hsu and Shengli Hu and Xin Hu and Joost Huizinga and Shantanu Jain and Shawn Jain and Joanne Jang and Angela Jiang and Roger Jiang and Haozhun Jin and Denny Jin and Shino Jomoto and Billie Jonn and Heewoo Jun and Tomer Kaftan and Łukasz Kaiser and Ali Kamali and Ingmar Kanitscheider and Nitish Shirish Keskar and Tabarak Khan and Logan Kilpatrick and Jong Wook Kim and Christina Kim and Yongjik Kim and Jan Hendrik Kirchner and Jamie Kiros and Matt Knight and Daniel Kokotajlo and Łukasz Kondraciuk and Andrew Kondrich and Aris Konstantinidis and Kyle Kosic and Gretchen Krueger and Vishal Kuo and Michael Lampe and Ikai Lan and Teddy Lee and Jan Leike and Jade Leung and Daniel Levy and Chak Ming Li and Rachel Lim and Molly Lin and Stephanie Lin and Mateusz Litwin and Theresa Lopez and Ryan Lowe and Patricia Lue and Anna Makanju and Kim Malfacini and Sam Manning and Todor Markov and Yaniv Markovski and Bianca Martin and Katie Mayer and Andrew Mayne and Bob McGrew and Scott Mayer McKinney and Christine McLeavey and Paul McMillan and Jake McNeil and David Medina and Aalok Mehta and Jacob Menick and Luke Metz and Andrey Mishchenko and Pamela Mishkin and Vinnie Monaco and Evan Morikawa and Daniel Mossing and Tong Mu and Mira Murati and Oleg Murk and David Mély and Ashvin Nair and Reiichiro Nakano and Rajeev Nayak and Arvind Neelakantan and Richard Ngo and Hyeonwoo Noh and Long Ouyang and Cullen O'Keefe and Jakub Pachocki and Alex Paino and Joe Palermo and Ashley Pantuliano and Giambattista Parascandolo and Joel Parish and Emy Parparita and Alex Passos and Mikhail Pavlov and Andrew Peng and Adam Perelman and Filipe de Avila Belbute Peres and Michael Petrov and Henrique Ponde de Oliveira Pinto and Michael and Pokorny and Michelle Pokrass and Vitchyr H. Pong and Tolly Powell and Alethea Power and Boris Power and Elizabeth Proehl and Raul Puri and Alec Radford and Jack Rae and Aditya Ramesh and Cameron Raymond and Francis Real and Kendra Rimbach and Carl Ross and Bob Rotsted and Henri Roussez and Nick Ryder and Mario Saltarelli and Ted Sanders and Shibani Santurkar and Girish Sastry and Heather Schmidt and David Schnurr and John Schulman and Daniel Selsam and Kyla Sheppard and Toki Sherbakov and Jessica Shieh and Sarah Shoker and Pranav Shyam and Szymon Sidor and Eric Sigler and Maddie Simens and Jordan Sitkin and Katarina Slama and Ian Sohl and Benjamin Sokolowsky and Yang Song and Natalie Staudacher and Felipe Petroski Such and Natalie Summers and Ilya Sutskever and Jie Tang and Nikolas Tezak and Madeleine B. Thompson and Phil Tillet and Amin Tootoonchian and Elizabeth Tseng and Preston Tuggle and Nick Turley and Jerry Tworek and Juan Felipe Cerón Uribe and Andrea Vallone and Arun Vijayvergiya and Chelsea Voss and Carroll Wainwright and Justin Jay Wang and Alvin Wang and Ben Wang and Jonathan Ward and Jason Wei and CJ Weinmann and Akila Welihinda and Peter Welinder and Jiayi Weng and Lilian Weng and Matt Wiethoff and Dave Willner and Clemens Winter and Samuel Wolrich and Hannah Wong and Lauren Workman and Sherwin Wu and Jeff Wu and Michael Wu and Kai Xiao and Tao Xu and Sarah Yoo and Kevin Yu and Qiming Yuan and Wojciech Zaremba and Rowan Zellers and Chong Zhang and Marvin Zhang and Shengjia Zhao and Tianhao Zheng and Juntang Zhuang and William Zhuk and Barret Zoph},
      year={2024},
      eprint={2303.08774},
      archivePrefix={arXiv},
      primaryClass={cs.CL},
      url={https://arxiv.org/abs/2303.08774}, 
}

@misc{grattafiori2024llama3herdmodels,
      title={The Llama 3 Herd of Models}, 
      author={Aaron Grattafiori and Abhimanyu Dubey and Abhinav Jauhri and Abhinav Pandey and Abhishek Kadian and Ahmad Al-Dahle and Aiesha Letman and Akhil Mathur and Alan Schelten and Alex Vaughan and Amy Yang and Angela Fan and Anirudh Goyal and Anthony Hartshorn and Aobo Yang and Archi Mitra and Archie Sravankumar and Artem Korenev and Arthur Hinsvark and Arun Rao and Aston Zhang and Aurelien Rodriguez and Austen Gregerson and Ava Spataru and Baptiste Roziere and Bethany Biron and Binh Tang and Bobbie Chern and Charlotte Caucheteux and Chaya Nayak and Chloe Bi and Chris Marra and Chris McConnell and Christian Keller and Christophe Touret and Chunyang Wu and Corinne Wong and Cristian Canton Ferrer and Cyrus Nikolaidis and Damien Allonsius and Daniel Song and Danielle Pintz and Danny Livshits and Danny Wyatt and David Esiobu and Dhruv Choudhary and Dhruv Mahajan and Diego Garcia-Olano and Diego Perino and Dieuwke Hupkes and Egor Lakomkin and Ehab AlBadawy and Elina Lobanova and Emily Dinan and Eric Michael Smith and Filip Radenovic and Francisco Guzmán and Frank Zhang and Gabriel Synnaeve and Gabrielle Lee and Georgia Lewis Anderson and Govind Thattai and Graeme Nail and Gregoire Mialon and Guan Pang and Guillem Cucurell and Hailey Nguyen and Hannah Korevaar and Hu Xu and Hugo Touvron and Iliyan Zarov and Imanol Arrieta Ibarra and Isabel Kloumann and Ishan Misra and Ivan Evtimov and Jack Zhang and Jade Copet and Jaewon Lee and Jan Geffert and Jana Vranes and Jason Park and Jay Mahadeokar and Jeet Shah and Jelmer van der Linde and Jennifer Billock and Jenny Hong and Jenya Lee and Jeremy Fu and Jianfeng Chi and Jianyu Huang and Jiawen Liu and Jie Wang and Jiecao Yu and Joanna Bitton and Joe Spisak and Jongsoo Park and Joseph Rocca and Joshua Johnstun and Joshua Saxe and Junteng Jia and Kalyan Vasuden Alwala and Karthik Prasad and Kartikeya Upasani and Kate Plawiak and Ke Li and Kenneth Heafield and Kevin Stone and Khalid El-Arini and Krithika Iyer and Kshitiz Malik and Kuenley Chiu and Kunal Bhalla and Kushal Lakhotia and Lauren Rantala-Yeary and Laurens van der Maaten and Lawrence Chen and Liang Tan and Liz Jenkins and Louis Martin and Lovish Madaan and Lubo Malo and Lukas Blecher and Lukas Landzaat and Luke de Oliveira and Madeline Muzzi and Mahesh Pasupuleti and Mannat Singh and Manohar Paluri and Marcin Kardas and Maria Tsimpoukelli and Mathew Oldham and Mathieu Rita and Maya Pavlova and Melanie Kambadur and Mike Lewis and Min Si and Mitesh Kumar Singh and Mona Hassan and Naman Goyal and Narjes Torabi and Nikolay Bashlykov and Nikolay Bogoychev and Niladri Chatterji and Ning Zhang and Olivier Duchenne and Onur Çelebi and Patrick Alrassy and Pengchuan Zhang and Pengwei Li and Petar Vasic and Peter Weng and Prajjwal Bhargava and Pratik Dubal and Praveen Krishnan and Punit Singh Koura and Puxin Xu and Qing He and Qingxiao Dong and Ragavan Srinivasan and Raj Ganapathy and Ramon Calderer and Ricardo Silveira Cabral and Robert Stojnic and Roberta Raileanu and Rohan Maheswari and Rohit Girdhar and Rohit Patel and Romain Sauvestre and Ronnie Polidoro and Roshan Sumbaly and Ross Taylor and Ruan Silva and Rui Hou and Rui Wang and Saghar Hosseini and Sahana Chennabasappa and Sanjay Singh and Sean Bell and Seohyun Sonia Kim and Sergey Edunov and Shaoliang Nie and Sharan Narang and Sharath Raparthy and Sheng Shen and Shengye Wan and Shruti Bhosale and Shun Zhang and Simon Vandenhende and Soumya Batra and Spencer Whitman and Sten Sootla and Stephane Collot and Suchin Gururangan and Sydney Borodinsky and Tamar Herman and Tara Fowler and Tarek Sheasha and Thomas Georgiou and Thomas Scialom and Tobias Speckbacher and Todor Mihaylov and Tong Xiao and Ujjwal Karn and Vedanuj Goswami and Vibhor Gupta and Vignesh Ramanathan and Viktor Kerkez and Vincent Gonguet and Virginie Do and Vish Vogeti and Vítor Albiero and Vladan Petrovic and Weiwei Chu and Wenhan Xiong and Wenyin Fu and Whitney Meers and Xavier Martinet and Xiaodong Wang and Xiaofang Wang and Xiaoqing Ellen Tan and Xide Xia and Xinfeng Xie and Xuchao Jia and Xuewei Wang and Yaelle Goldschlag and Yashesh Gaur and Yasmine Babaei and Yi Wen and Yiwen Song and Yuchen Zhang and Yue Li and Yuning Mao and Zacharie Delpierre Coudert and Zheng Yan and Zhengxing Chen and Zoe Papakipos and Aaditya Singh and Aayushi Srivastava and Abha Jain and Adam Kelsey and Adam Shajnfeld and Adithya Gangidi and Adolfo Victoria and Ahuva Goldstand and Ajay Menon and Ajay Sharma and Alex Boesenberg and Alexei Baevski and Allie Feinstein and Amanda Kallet and Amit Sangani and Amos Teo and Anam Yunus and Andrei Lupu and Andres Alvarado and Andrew Caples and Andrew Gu and Andrew Ho and Andrew Poulton and Andrew Ryan and Ankit Ramchandani and Annie Dong and Annie Franco and Anuj Goyal and Aparajita Saraf and Arkabandhu Chowdhury and Ashley Gabriel and Ashwin Bharambe and Assaf Eisenman and Azadeh Yazdan and Beau James and Ben Maurer and Benjamin Leonhardi and Bernie Huang and Beth Loyd and Beto De Paola and Bhargavi Paranjape and Bing Liu and Bo Wu and Boyu Ni and Braden Hancock and Bram Wasti and Brandon Spence and Brani Stojkovic and Brian Gamido and Britt Montalvo and Carl Parker and Carly Burton and Catalina Mejia and Ce Liu and Changhan Wang and Changkyu Kim and Chao Zhou and Chester Hu and Ching-Hsiang Chu and Chris Cai and Chris Tindal and Christoph Feichtenhofer and Cynthia Gao and Damon Civin and Dana Beaty and Daniel Kreymer and Daniel Li and David Adkins and David Xu and Davide Testuggine and Delia David and Devi Parikh and Diana Liskovich and Didem Foss and Dingkang Wang and Duc Le and Dustin Holland and Edward Dowling and Eissa Jamil and Elaine Montgomery and Eleonora Presani and Emily Hahn and Emily Wood and Eric-Tuan Le and Erik Brinkman and Esteban Arcaute and Evan Dunbar and Evan Smothers and Fei Sun and Felix Kreuk and Feng Tian and Filippos Kokkinos and Firat Ozgenel and Francesco Caggioni and Frank Kanayet and Frank Seide and Gabriela Medina Florez and Gabriella Schwarz and Gada Badeer and Georgia Swee and Gil Halpern and Grant Herman and Grigory Sizov and Guangyi and Zhang and Guna Lakshminarayanan and Hakan Inan and Hamid Shojanazeri and Han Zou and Hannah Wang and Hanwen Zha and Haroun Habeeb and Harrison Rudolph and Helen Suk and Henry Aspegren and Hunter Goldman and Hongyuan Zhan and Ibrahim Damlaj and Igor Molybog and Igor Tufanov and Ilias Leontiadis and Irina-Elena Veliche and Itai Gat and Jake Weissman and James Geboski and James Kohli and Janice Lam and Japhet Asher and Jean-Baptiste Gaya and Jeff Marcus and Jeff Tang and Jennifer Chan and Jenny Zhen and Jeremy Reizenstein and Jeremy Teboul and Jessica Zhong and Jian Jin and Jingyi Yang and Joe Cummings and Jon Carvill and Jon Shepard and Jonathan McPhie and Jonathan Torres and Josh Ginsburg and Junjie Wang and Kai Wu and Kam Hou U and Karan Saxena and Kartikay Khandelwal and Katayoun Zand and Kathy Matosich and Kaushik Veeraraghavan and Kelly Michelena and Keqian Li and Kiran Jagadeesh and Kun Huang and Kunal Chawla and Kyle Huang and Lailin Chen and Lakshya Garg and Lavender A and Leandro Silva and Lee Bell and Lei Zhang and Liangpeng Guo and Licheng Yu and Liron Moshkovich and Luca Wehrstedt and Madian Khabsa and Manav Avalani and Manish Bhatt and Martynas Mankus and Matan Hasson and Matthew Lennie and Matthias Reso and Maxim Groshev and Maxim Naumov and Maya Lathi and Meghan Keneally and Miao Liu and Michael L. Seltzer and Michal Valko and Michelle Restrepo and Mihir Patel and Mik Vyatskov and Mikayel Samvelyan and Mike Clark and Mike Macey and Mike Wang and Miquel Jubert Hermoso and Mo Metanat and Mohammad Rastegari and Munish Bansal and Nandhini Santhanam and Natascha Parks and Natasha White and Navyata Bawa and Nayan Singhal and Nick Egebo and Nicolas Usunier and Nikhil Mehta and Nikolay Pavlovich Laptev and Ning Dong and Norman Cheng and Oleg Chernoguz and Olivia Hart and Omkar Salpekar and Ozlem Kalinli and Parkin Kent and Parth Parekh and Paul Saab and Pavan Balaji and Pedro Rittner and Philip Bontrager and Pierre Roux and Piotr Dollar and Polina Zvyagina and Prashant Ratanchandani and Pritish Yuvraj and Qian Liang and Rachad Alao and Rachel Rodriguez and Rafi Ayub and Raghotham Murthy and Raghu Nayani and Rahul Mitra and Rangaprabhu Parthasarathy and Raymond Li and Rebekkah Hogan and Robin Battey and Rocky Wang and Russ Howes and Ruty Rinott and Sachin Mehta and Sachin Siby and Sai Jayesh Bondu and Samyak Datta and Sara Chugh and Sara Hunt and Sargun Dhillon and Sasha Sidorov and Satadru Pan and Saurabh Mahajan and Saurabh Verma and Seiji Yamamoto and Sharadh Ramaswamy and Shaun Lindsay and Shaun Lindsay and Sheng Feng and Shenghao Lin and Shengxin Cindy Zha and Shishir Patil and Shiva Shankar and Shuqiang Zhang and Shuqiang Zhang and Sinong Wang and Sneha Agarwal and Soji Sajuyigbe and Soumith Chintala and Stephanie Max and Stephen Chen and Steve Kehoe and Steve Satterfield and Sudarshan Govindaprasad and Sumit Gupta and Summer Deng and Sungmin Cho and Sunny Virk and Suraj Subramanian and Sy Choudhury and Sydney Goldman and Tal Remez and Tamar Glaser and Tamara Best and Thilo Koehler and Thomas Robinson and Tianhe Li and Tianjun Zhang and Tim Matthews and Timothy Chou and Tzook Shaked and Varun Vontimitta and Victoria Ajayi and Victoria Montanez and Vijai Mohan and Vinay Satish Kumar and Vishal Mangla and Vlad Ionescu and Vlad Poenaru and Vlad Tiberiu Mihailescu and Vladimir Ivanov and Wei Li and Wenchen Wang and Wenwen Jiang and Wes Bouaziz and Will Constable and Xiaocheng Tang and Xiaojian Wu and Xiaolan Wang and Xilun Wu and Xinbo Gao and Yaniv Kleinman and Yanjun Chen and Ye Hu and Ye Jia and Ye Qi and Yenda Li and Yilin Zhang and Ying Zhang and Yossi Adi and Youngjin Nam and Yu and Wang and Yu Zhao and Yuchen Hao and Yundi Qian and Yunlu Li and Yuzi He and Zach Rait and Zachary DeVito and Zef Rosnbrick and Zhaoduo Wen and Zhenyu Yang and Zhiwei Zhao and Zhiyu Ma},
      year={2024},
      eprint={2407.21783},
      archivePrefix={arXiv},
      primaryClass={cs.AI},
      url={https://arxiv.org/abs/2407.21783}, 
}

@misc{touvron2023llama2openfoundation,
      title={Llama 2: Open Foundation and Fine-Tuned Chat Models}, 
      author={Hugo Touvron and Louis Martin and Kevin Stone and Peter Albert and Amjad Almahairi and Yasmine Babaei and Nikolay Bashlykov and Soumya Batra and Prajjwal Bhargava and Shruti Bhosale and Dan Bikel and Lukas Blecher and Cristian Canton Ferrer and Moya Chen and Guillem Cucurull and David Esiobu and Jude Fernandes and Jeremy Fu and Wenyin Fu and Brian Fuller and Cynthia Gao and Vedanuj Goswami and Naman Goyal and Anthony Hartshorn and Saghar Hosseini and Rui Hou and Hakan Inan and Marcin Kardas and Viktor Kerkez and Madian Khabsa and Isabel Kloumann and Artem Korenev and Punit Singh Koura and Marie-Anne Lachaux and Thibaut Lavril and Jenya Lee and Diana Liskovich and Yinghai Lu and Yuning Mao and Xavier Martinet and Todor Mihaylov and Pushkar Mishra and Igor Molybog and Yixin Nie and Andrew Poulton and Jeremy Reizenstein and Rashi Rungta and Kalyan Saladi and Alan Schelten and Ruan Silva and Eric Michael Smith and Ranjan Subramanian and Xiaoqing Ellen Tan and Binh Tang and Ross Taylor and Adina Williams and Jian Xiang Kuan and Puxin Xu and Zheng Yan and Iliyan Zarov and Yuchen Zhang and Angela Fan and Melanie Kambadur and Sharan Narang and Aurelien Rodriguez and Robert Stojnic and Sergey Edunov and Thomas Scialom},
      year={2023},
      eprint={2307.09288},
      archivePrefix={arXiv},
      primaryClass={cs.CL},
      url={https://arxiv.org/abs/2307.09288}, 
}

@misc{deepseekai2025deepseekv3technicalreport,
      title={DeepSeek-V3 Technical Report}, 
      author={DeepSeek-AI and Aixin Liu and Bei Feng and Bing Xue and Bingxuan Wang and Bochao Wu and Chengda Lu and Chenggang Zhao and Chengqi Deng and Chenyu Zhang and Chong Ruan and Damai Dai and Daya Guo and Dejian Yang and Deli Chen and Dongjie Ji and Erhang Li and Fangyun Lin and Fucong Dai and Fuli Luo and Guangbo Hao and Guanting Chen and Guowei Li and H. Zhang and Han Bao and Hanwei Xu and Haocheng Wang and Haowei Zhang and Honghui Ding and Huajian Xin and Huazuo Gao and Hui Li and Hui Qu and J. L. Cai and Jian Liang and Jianzhong Guo and Jiaqi Ni and Jiashi Li and Jiawei Wang and Jin Chen and Jingchang Chen and Jingyang Yuan and Junjie Qiu and Junlong Li and Junxiao Song and Kai Dong and Kai Hu and Kaige Gao and Kang Guan and Kexin Huang and Kuai Yu and Lean Wang and Lecong Zhang and Lei Xu and Leyi Xia and Liang Zhao and Litong Wang and Liyue Zhang and Meng Li and Miaojun Wang and Mingchuan Zhang and Minghua Zhang and Minghui Tang and Mingming Li and Ning Tian and Panpan Huang and Peiyi Wang and Peng Zhang and Qiancheng Wang and Qihao Zhu and Qinyu Chen and Qiushi Du and R. J. Chen and R. L. Jin and Ruiqi Ge and Ruisong Zhang and Ruizhe Pan and Runji Wang and Runxin Xu and Ruoyu Zhang and Ruyi Chen and S. S. Li and Shanghao Lu and Shangyan Zhou and Shanhuang Chen and Shaoqing Wu and Shengfeng Ye and Shengfeng Ye and Shirong Ma and Shiyu Wang and Shuang Zhou and Shuiping Yu and Shunfeng Zhou and Shuting Pan and T. Wang and Tao Yun and Tian Pei and Tianyu Sun and W. L. Xiao and Wangding Zeng and Wanjia Zhao and Wei An and Wen Liu and Wenfeng Liang and Wenjun Gao and Wenqin Yu and Wentao Zhang and X. Q. Li and Xiangyue Jin and Xianzu Wang and Xiao Bi and Xiaodong Liu and Xiaohan Wang and Xiaojin Shen and Xiaokang Chen and Xiaokang Zhang and Xiaosha Chen and Xiaotao Nie and Xiaowen Sun and Xiaoxiang Wang and Xin Cheng and Xin Liu and Xin Xie and Xingchao Liu and Xingkai Yu and Xinnan Song and Xinxia Shan and Xinyi Zhou and Xinyu Yang and Xinyuan Li and Xuecheng Su and Xuheng Lin and Y. K. Li and Y. Q. Wang and Y. X. Wei and Y. X. Zhu and Yang Zhang and Yanhong Xu and Yanhong Xu and Yanping Huang and Yao Li and Yao Zhao and Yaofeng Sun and Yaohui Li and Yaohui Wang and Yi Yu and Yi Zheng and Yichao Zhang and Yifan Shi and Yiliang Xiong and Ying He and Ying Tang and Yishi Piao and Yisong Wang and Yixuan Tan and Yiyang Ma and Yiyuan Liu and Yongqiang Guo and Yu Wu and Yuan Ou and Yuchen Zhu and Yuduan Wang and Yue Gong and Yuheng Zou and Yujia He and Yukun Zha and Yunfan Xiong and Yunxian Ma and Yuting Yan and Yuxiang Luo and Yuxiang You and Yuxuan Liu and Yuyang Zhou and Z. F. Wu and Z. Z. Ren and Zehui Ren and Zhangli Sha and Zhe Fu and Zhean Xu and Zhen Huang and Zhen Zhang and Zhenda Xie and Zhengyan Zhang and Zhewen Hao and Zhibin Gou and Zhicheng Ma and Zhigang Yan and Zhihong Shao and Zhipeng Xu and Zhiyu Wu and Zhongyu Zhang and Zhuoshu Li and Zihui Gu and Zijia Zhu and Zijun Liu and Zilin Li and Ziwei Xie and Ziyang Song and Ziyi Gao and Zizheng Pan},
      year={2025},
      eprint={2412.19437},
      archivePrefix={arXiv},
      primaryClass={cs.CL},
      url={https://arxiv.org/abs/2412.19437}, 
}

@misc{deepseekai2025deepseekr1incentivizingreasoningcapability,
      title={DeepSeek-R1: Incentivizing Reasoning Capability in LLMs via Reinforcement Learning}, 
      author={DeepSeek-AI and Daya Guo and Dejian Yang and Haowei Zhang and Junxiao Song and Ruoyu Zhang and Runxin Xu and Qihao Zhu and Shirong Ma and Peiyi Wang and Xiao Bi and Xiaokang Zhang and Xingkai Yu and Yu Wu and Z. F. Wu and Zhibin Gou and Zhihong Shao and Zhuoshu Li and Ziyi Gao and Aixin Liu and Bing Xue and Bingxuan Wang and Bochao Wu and Bei Feng and Chengda Lu and Chenggang Zhao and Chengqi Deng and Chenyu Zhang and Chong Ruan and Damai Dai and Deli Chen and Dongjie Ji and Erhang Li and Fangyun Lin and Fucong Dai and Fuli Luo and Guangbo Hao and Guanting Chen and Guowei Li and H. Zhang and Han Bao and Hanwei Xu and Haocheng Wang and Honghui Ding and Huajian Xin and Huazuo Gao and Hui Qu and Hui Li and Jianzhong Guo and Jiashi Li and Jiawei Wang and Jingchang Chen and Jingyang Yuan and Junjie Qiu and Junlong Li and J. L. Cai and Jiaqi Ni and Jian Liang and Jin Chen and Kai Dong and Kai Hu and Kaige Gao and Kang Guan and Kexin Huang and Kuai Yu and Lean Wang and Lecong Zhang and Liang Zhao and Litong Wang and Liyue Zhang and Lei Xu and Leyi Xia and Mingchuan Zhang and Minghua Zhang and Minghui Tang and Meng Li and Miaojun Wang and Mingming Li and Ning Tian and Panpan Huang and Peng Zhang and Qiancheng Wang and Qinyu Chen and Qiushi Du and Ruiqi Ge and Ruisong Zhang and Ruizhe Pan and Runji Wang and R. J. Chen and R. L. Jin and Ruyi Chen and Shanghao Lu and Shangyan Zhou and Shanhuang Chen and Shengfeng Ye and Shiyu Wang and Shuiping Yu and Shunfeng Zhou and Shuting Pan and S. S. Li and Shuang Zhou and Shaoqing Wu and Shengfeng Ye and Tao Yun and Tian Pei and Tianyu Sun and T. Wang and Wangding Zeng and Wanjia Zhao and Wen Liu and Wenfeng Liang and Wenjun Gao and Wenqin Yu and Wentao Zhang and W. L. Xiao and Wei An and Xiaodong Liu and Xiaohan Wang and Xiaokang Chen and Xiaotao Nie and Xin Cheng and Xin Liu and Xin Xie and Xingchao Liu and Xinyu Yang and Xinyuan Li and Xuecheng Su and Xuheng Lin and X. Q. Li and Xiangyue Jin and Xiaojin Shen and Xiaosha Chen and Xiaowen Sun and Xiaoxiang Wang and Xinnan Song and Xinyi Zhou and Xianzu Wang and Xinxia Shan and Y. K. Li and Y. Q. Wang and Y. X. Wei and Yang Zhang and Yanhong Xu and Yao Li and Yao Zhao and Yaofeng Sun and Yaohui Wang and Yi Yu and Yichao Zhang and Yifan Shi and Yiliang Xiong and Ying He and Yishi Piao and Yisong Wang and Yixuan Tan and Yiyang Ma and Yiyuan Liu and Yongqiang Guo and Yuan Ou and Yuduan Wang and Yue Gong and Yuheng Zou and Yujia He and Yunfan Xiong and Yuxiang Luo and Yuxiang You and Yuxuan Liu and Yuyang Zhou and Y. X. Zhu and Yanhong Xu and Yanping Huang and Yaohui Li and Yi Zheng and Yuchen Zhu and Yunxian Ma and Ying Tang and Yukun Zha and Yuting Yan and Z. Z. Ren and Zehui Ren and Zhangli Sha and Zhe Fu and Zhean Xu and Zhenda Xie and Zhengyan Zhang and Zhewen Hao and Zhicheng Ma and Zhigang Yan and Zhiyu Wu and Zihui Gu and Zijia Zhu and Zijun Liu and Zilin Li and Ziwei Xie and Ziyang Song and Zizheng Pan and Zhen Huang and Zhipeng Xu and Zhongyu Zhang and Zhen Zhang},
      year={2025},
      eprint={2501.12948},
      archivePrefix={arXiv},
      primaryClass={cs.CL},
      url={https://arxiv.org/abs/2501.12948}, 
}

@inproceedings{NEURIPS2023_949f0f8f,
 author = {K\"{o}pf, Andreas and Kilcher, Yannic and von R\"{u}tte, Dimitri and Anagnostidis, Sotiris and Tam, Zhi Rui and Stevens, Keith and Barhoum, Abdullah and Nguyen, Duc and Stanley, Oliver and Nagyfi, Rich\'{a}rd and ES, Shahul and Suri, Sameer and Glushkov, David and Dantuluri, Arnav and Maguire, Andrew and Schuhmann, Christoph and Nguyen, Huu and Mattick, Alexander},
 booktitle = {Advances in Neural Information Processing Systems},
 editor = {A. Oh and T. Naumann and A. Globerson and K. Saenko and M. Hardt and S. Levine},
 pages = {47669--47681},
 publisher = {Curran Associates, Inc.},
 title = {OpenAssistant Conversations - Democratizing Large Language Model Alignment},
 url = {https://proceedings.neurips.cc/paper_files/paper/2023/file/949f0f8f32267d297c2d4e3ee10a2e7e-Paper-Datasets_and_Benchmarks.pdf},
 volume = {36},
 year = {2023}
}

@article{pearson1895regression,
  author = {Pearson, Karl},
  title = {Notes on regression and inheritance in the case of two parents},
  journal = {Proceedings of the Royal Society of London},
  volume = {58},
  pages = {240--242},
  year = {1895},
  doi = {10.1098/rspl.1895.0041}
}

@inproceedings{Rasley_Rajbhandari_Ruwase_He_2020,
  title={Deepspeed: System optimizations enable training deep learning models with over 100 billion parameters},
  author={Rasley, Jeff and Rajbhandari, Samyam and Ruwase, Olatunji and He, Yuxiong},
  booktitle={Proceedings of the ACM SIGKDD International Conference on Knowledge Discovery \& Data Mining},
  pages={3505--3506},
  year={2020}
}

@inproceedings{NEURIPS2020_1f89885d,
 author = {Stiennon, Nisan and Ouyang, Long and Wu, Jeffrey and Ziegler, Daniel and Lowe, Ryan and Voss, Chelsea and Radford, Alec and Amodei, Dario and Christiano, Paul F},
 booktitle = {Advances in Neural Information Processing Systems},
 editor = {H. Larochelle and M. Ranzato and R. Hadsell and M.F. Balcan and H. Lin},
 pages = {3008--3021},
 publisher = {Curran Associates, Inc.},
 title = {Learning to summarize with human feedback},
 url = {https://proceedings.neurips.cc/paper_files/paper/2020/file/1f89885d556929e98d3ef9b86448f951-Paper.pdf},
 volume = {33},
 year = {2020}
}

@inproceedings{
eisenstein2024helping,
title={Helping or Herding? Reward Model Ensembles Mitigate but do not Eliminate Reward Hacking},
author={Jacob Eisenstein and Chirag Nagpal and Alekh Agarwal and Ahmad Beirami and Alexander Nicholas D'Amour and Krishnamurthy Dj Dvijotham and Adam Fisch and Katherine A Heller and Stephen Robert Pfohl and Deepak Ramachandran and Peter Shaw and Jonathan Berant},
booktitle={First Conference on Language Modeling},
year={2024},
url={https://openreview.net/forum?id=5u1GpUkKtG}
}

@inproceedings{
shen2023loose,
title={Loose lips sink ships: Mitigating Length Bias in Reinforcement Learning from Human Feedback},
author={Wei Shen and Rui Zheng and Wenyu Zhan and Jun Zhao and Shihan Dou and Tao Gui and Qi Zhang and Xuanjing Huang},
booktitle={The 2023 Conference on Empirical Methods in Natural Language Processing},
year={2023},
url={https://openreview.net/forum?id=qq6ctdUwCX}
}

@inproceedings{
huang2025posthoc,
title={Post-hoc Reward Calibration: A Case Study on Length Bias},
author={Zeyu Huang and Zihan Qiu and Zili Wang and Edoardo Ponti and Ivan Titov},
booktitle={The Thirteenth International Conference on Learning Representations},
year={2025},
url={https://openreview.net/forum?id=Iu8RytBaji}
}

@inproceedings{pang2023reward,
  title={Reward Gaming in Conditional Text Generation},
  author={Pang, Richard Yuanzhe and Padmakumar, Vishakh and Sellam, Thibault and Parikh, Ankur and He, He},
  booktitle={Proceedings of the 61st Annual Meeting of the Association for Computational Linguistics (Volume 1: Long Papers)},
  pages={4746--4763},
  year={2023}
}

@article{lambert2023alignment,
  title={The alignment ceiling: Objective mismatch in reinforcement learning from human feedback},
  author={Lambert, Nathan and Calandra, Roberto},
  journal={arXiv preprint arXiv:2311.00168},
  year={2023}
}

@misc{dong2024rlhfworkflowrewardmodeling,
      title={RLHF Workflow: From Reward Modeling to Online RLHF}, 
      author={Hanze Dong and Wei Xiong and Bo Pang and Haoxiang Wang and Han Zhao and Yingbo Zhou and Nan Jiang and Doyen Sahoo and Caiming Xiong and Tong Zhang},
      year={2024},
      eprint={2405.07863},
      archivePrefix={arXiv},
      primaryClass={cs.LG},
      url={https://arxiv.org/abs/2405.07863}, 
}

@misc{chen2024noisecontrastivealignmentlanguage,
      title={Noise Contrastive Alignment of Language Models with Explicit Rewards}, 
      author={Huayu Chen and Guande He and Lifan Yuan and Ganqu Cui and Hang Su and Jun Zhu},
      year={2024},
      eprint={2402.05369},
      archivePrefix={arXiv},
      primaryClass={cs.LG},
      url={https://arxiv.org/abs/2402.05369}, 
}

@inproceedings{hong-etal-2024-orpo,
    title = "{ORPO}: Monolithic Preference Optimization without Reference Model",
    author = "Hong, Jiwoo  and
      Lee, Noah  and
      Thorne, James",
    editor = "Al-Onaizan, Yaser  and
      Bansal, Mohit  and
      Chen, Yun-Nung",
    booktitle = "Proceedings of the 2024 Conference on Empirical Methods in Natural Language Processing",
    month = nov,
    year = "2024",
    address = "Miami, Florida, USA",
    publisher = "Association for Computational Linguistics",
    url = "https://aclanthology.org/2024.emnlp-main.626/",
    doi = "10.18653/v1/2024.emnlp-main.626",
    pages = "11170--11189",
   
}

@InProceedings{pmlr-v238-gheshlaghi-azar24a,
  title = 	 {A General Theoretical Paradigm to Understand Learning from Human Preferences},
  author =       {Gheshlaghi Azar, Mohammad and Daniel Guo, Zhaohan and Piot, Bilal and Munos, Remi and Rowland, Mark and Valko, Michal and Calandriello, Daniele},
  booktitle = 	 {Proceedings of The 27th International Conference on Artificial Intelligence and Statistics},
  pages = 	 {4447--4455},
  year = 	 {2024},
  editor = 	 {Dasgupta, Sanjoy and Mandt, Stephan and Li, Yingzhen},
  volume = 	 {238},
  series = 	 {Proceedings of Machine Learning Research},
  month = 	 {02--04 May},
  publisher =    {PMLR},
  url = 	 {https://proceedings.mlr.press/v238/gheshlaghi-azar24a.html},
  
}

@article{ethayarajh2024kto,
  title={Kto: Model alignment as prospect theoretic optimization},
  author={Ethayarajh, Kawin and Xu, Winnie and Muennighoff, Niklas and Jurafsky, Dan and Kiela, Douwe},
  journal={arXiv preprint arXiv:2402.01306},
  year={2024}
}

@misc{richemond2024offlineregularisedreinforcementlearning,
      title={Offline Regularised Reinforcement Learning for Large Language Models Alignment}, 
      author={Pierre Harvey Richemond and Yunhao Tang and Daniel Guo and Daniele Calandriello and Mohammad Gheshlaghi Azar and Rafael Rafailov and Bernardo Avila Pires and Eugene Tarassov and Lucas Spangher and Will Ellsworth and Aliaksei Severyn and Jonathan Mallinson and Lior Shani and Gil Shamir and Rishabh Joshi and Tianqi Liu and Remi Munos and Bilal Piot},
      year={2024},
      eprint={2405.19107},
      archivePrefix={arXiv},
      primaryClass={cs.LG},
      url={https://arxiv.org/abs/2405.19107}, 
}

@misc{gemmateam2024gemma2improvingopen,
      title={Gemma 2: Improving Open Language Models at a Practical Size}, 
      author={Gemma Team and Morgane Riviere and Shreya Pathak and Pier Giuseppe Sessa and Cassidy Hardin and Surya Bhupatiraju and Léonard Hussenot and Thomas Mesnard and Bobak Shahriari and Alexandre Ramé and Johan Ferret and Peter Liu and Pouya Tafti and Abe Friesen and Michelle Casbon and Sabela Ramos and Ravin Kumar and Charline Le Lan and Sammy Jerome and Anton Tsitsulin and Nino Vieillard and Piotr Stanczyk and Sertan Girgin and Nikola Momchev and Matt Hoffman and Shantanu Thakoor and Jean-Bastien Grill and Behnam Neyshabur and Olivier Bachem and Alanna Walton and Aliaksei Severyn and Alicia Parrish and Aliya Ahmad and Allen Hutchison and Alvin Abdagic and Amanda Carl and Amy Shen and Andy Brock and Andy Coenen and Anthony Laforge and Antonia Paterson and Ben Bastian and Bilal Piot and Bo Wu and Brandon Royal and Charlie Chen and Chintu Kumar and Chris Perry and Chris Welty and Christopher A. Choquette-Choo and Danila Sinopalnikov and David Weinberger and Dimple Vijaykumar and Dominika Rogozińska and Dustin Herbison and Elisa Bandy and Emma Wang and Eric Noland and Erica Moreira and Evan Senter and Evgenii Eltyshev and Francesco Visin and Gabriel Rasskin and Gary Wei and Glenn Cameron and Gus Martins and Hadi Hashemi and Hanna Klimczak-Plucińska and Harleen Batra and Harsh Dhand and Ivan Nardini and Jacinda Mein and Jack Zhou and James Svensson and Jeff Stanway and Jetha Chan and Jin Peng Zhou and Joana Carrasqueira and Joana Iljazi and Jocelyn Becker and Joe Fernandez and Joost van Amersfoort and Josh Gordon and Josh Lipschultz and Josh Newlan and Ju-yeong Ji and Kareem Mohamed and Kartikeya Badola and Kat Black and Katie Millican and Keelin McDonell and Kelvin Nguyen and Kiranbir Sodhia and Kish Greene and Lars Lowe Sjoesund and Lauren Usui and Laurent Sifre and Lena Heuermann and Leticia Lago and Lilly McNealus and Livio Baldini Soares and Logan Kilpatrick and Lucas Dixon and Luciano Martins and Machel Reid and Manvinder Singh and Mark Iverson and Martin Görner and Mat Velloso and Mateo Wirth and Matt Davidow and Matt Miller and Matthew Rahtz and Matthew Watson and Meg Risdal and Mehran Kazemi and Michael Moynihan and Ming Zhang and Minsuk Kahng and Minwoo Park and Mofi Rahman and Mohit Khatwani and Natalie Dao and Nenshad Bardoliwalla and Nesh Devanathan and Neta Dumai and Nilay Chauhan and Oscar Wahltinez and Pankil Botarda and Parker Barnes and Paul Barham and Paul Michel and Pengchong Jin and Petko Georgiev and Phil Culliton and Pradeep Kuppala and Ramona Comanescu and Ramona Merhej and Reena Jana and Reza Ardeshir Rokni and Rishabh Agarwal and Ryan Mullins and Samaneh Saadat and Sara Mc Carthy and Sarah Cogan and Sarah Perrin and Sébastien M. R. Arnold and Sebastian Krause and Shengyang Dai and Shruti Garg and Shruti Sheth and Sue Ronstrom and Susan Chan and Timothy Jordan and Ting Yu and Tom Eccles and Tom Hennigan and Tomas Kocisky and Tulsee Doshi and Vihan Jain and Vikas Yadav and Vilobh Meshram and Vishal Dharmadhikari and Warren Barkley and Wei Wei and Wenming Ye and Woohyun Han and Woosuk Kwon and Xiang Xu and Zhe Shen and Zhitao Gong and Zichuan Wei and Victor Cotruta and Phoebe Kirk and Anand Rao and Minh Giang and Ludovic Peran and Tris Warkentin and Eli Collins and Joelle Barral and Zoubin Ghahramani and Raia Hadsell and D. Sculley and Jeanine Banks and Anca Dragan and Slav Petrov and Oriol Vinyals and Jeff Dean and Demis Hassabis and Koray Kavukcuoglu and Clement Farabet and Elena Buchatskaya and Sebastian Borgeaud and Noah Fiedel and Armand Joulin and Kathleen Kenealy and Robert Dadashi and Alek Andreev},
      year={2024},
      eprint={2408.00118},
      archivePrefix={arXiv},
      primaryClass={cs.CL},
      url={https://arxiv.org/abs/2408.00118}, 
}

@inproceedings{zheng2023judging,
  title={Judging LLM-as-a-judge with MT-bench and Chatbot Arena},
  author={Lianmin Zheng and Wei-Lin Chiang and Ying Sheng and Siyuan Zhuang and Zhanghao Wu and Yonghao Zhuang and Zi Lin and Zhuohan Li and Dacheng Li and Eric P. Xing and Hao Zhang and Joseph E. Gonzalez and Ion Stoica},
  booktitle={Proceedings of the 37th International Conference on Neural Information Processing Systems},
  pages={46595--46623},
  year={2023}
}

@article{IFEVAL,
  publtype={informal},
  author={Jeffrey Zhou and Tianjian Lu and Swaroop Mishra and Siddhartha Brahma and Sujoy Basu and Yi Luan and Denny Zhou and Le Hou},
  title={Instruction-Following Evaluation for Large Language Models},
  year={2023},
  cdate={1672531200000},
  journal={CoRR},
  volume={abs/2311.07911},
  url={https://doi.org/10.48550/arXiv.2311.07911}
}

@inproceedings{park-etal-2024-disentangling,
    title = "Disentangling Length from Quality in Direct Preference Optimization",
    author = "Park, Ryan  and
      Rafailov, Rafael  and
      Ermon, Stefano  and
      Finn, Chelsea",
    editor = "Ku, Lun-Wei  and
      Martins, Andre  and
      Srikumar, Vivek",
    booktitle = "Findings of the Association for Computational Linguistics: ACL 2024",
    month = aug,
    year = "2024",
    address = "Bangkok, Thailand",
    publisher = "Association for Computational Linguistics",
    url = "https://aclanthology.org/2024.findings-acl.297/",
    doi = "10.18653/v1/2024.findings-acl.297",
    pages = "4998--5017",

}

@inproceedings{lu-etal-2024-eliminating,
    title = "Eliminating Biased Length Reliance of Direct Preference Optimization via Down-Sampled {KL} Divergence",
    author = "Lu, Junru  and
      Li, Jiazheng  and
      An, Siyu  and
      Zhao, Meng  and
      He, Yulan  and
      Yin, Di  and
      Sun, Xing",
    editor = "Al-Onaizan, Yaser  and
      Bansal, Mohit  and
      Chen, Yun-Nung",
    booktitle = "Proceedings of the 2024 Conference on Empirical Methods in Natural Language Processing",
    month = nov,
    year = "2024",
    address = "Miami, Florida, USA",
    publisher = "Association for Computational Linguistics",
    url = "https://aclanthology.org/2024.emnlp-main.60/",
    doi = "10.18653/v1/2024.emnlp-main.60",
    pages = "1047--1067",

}

@inproceedings{bu-etal-2025-beyond,
    title = "Beyond Excess and Deficiency: Adaptive Length Bias Mitigation in Reward Models for {RLHF}",
    author = "Bu, Yuyan  and
      Huo, Liangyu  and
      Jing, Yi  and
      Yang, Qing",
    editor = "Chiruzzo, Luis  and
      Ritter, Alan  and
      Wang, Lu",
    booktitle = "Findings of the Association for Computational Linguistics: NAACL 2025",
    month = apr,
    year = "2025",
    address = "Albuquerque, New Mexico",
    publisher = "Association for Computational Linguistics",
    url = "https://aclanthology.org/2025.findings-naacl.169/",
    doi = "10.18653/v1/2025.findings-naacl.169",
    pages = "3091--3098",
    ISBN = "979-8-89176-195-7",
   
}
